%% file: main.tex
\documentclass[smallextended]{article}

\input{0_packages}

\begin{document}

\input{0_title_page}

\input{1_abstract}
\input{2_introduction}
\input{3_related_work}
\input{4_method}

\input{5_results}

\input{6_discussion}

\input{7_conclusion}

\input{8_acknowledgment}

\input{9_bibliography}

\end{document}

%% file: 0_packages.tex
\makeatletter
\def\maxwidth{ %
  \ifdim\Gin@nat@width>\linewidth
    \linewidth
  \else
    \Gin@nat@width
  \fi
}
\makeatother

\usepackage{array}
\usepackage{float}
\usepackage{lineno}
\usepackage{amsmath}
\usepackage{amssymb}
\usepackage{anysize}
\usepackage{fancyhdr}
\usepackage{multirow}
\usepackage{listings}
\usepackage{latexsym}
\usepackage[resetlabels,labeled]{multibib}
\usepackage{breakcites}

\usepackage{setspace}
\usepackage{pdflscape}
\usepackage{longtable}
\usepackage{tabularx}
\usepackage{booktabs}   
\usepackage{ltablex}
\usepackage{makecell}
\usepackage[demo]{graphicx}
\usepackage[utf8]{inputenc}
\usepackage{authblk}
\usepackage[caption = false]{subfig}
\usepackage[labeled,resetlabels]{multibib}
\usepackage[colorlinks=true,linkcolor=blue,citecolor=blue,allcolors=blue]{hyperref}
\usepackage[a4paper, left=15mm, right=15mm, top=20mm, bottom=20mm, includeheadfoot]{geometry}

\newcites{S}{Primary Studies}

%% file: 0_title_page.tex
\title{Knowledge Graphs in Manufacturing and Production: A Systematic Literature Review}
\author[*]{Georg Buchgeher}
\author[*]{David Gabauer}
\author[*]{Jorge Martinez-Gil}
\author[*,**]{Lisa Ehrlinger}
\affil[*]{Software Competence Center Hagenberg, Softwarepark 21, 4232 Hagenberg, Austria}
\affil[**]{Johannes Kepler University Linz, Altenberger Straße 69, 4040 Linz, Austria}

\date{}

\maketitle

%% file: 1_abstract.tex
%\begin{abstract}
%\noindent \textit{Context:} Knowledge graphs in manufacturing and production make production lines more efficient and flexible with a higher quality output. This combination makes knowledge graphs attractive for companies and supports to reach the industry 4.0 standards.\\
%\textit{Objective:} The objective of this paper is to carry out a systematic literature review of the field of knowledge graph in manufacturing and production in an attempt to characterize the state-of-the-art of this field identifying gaps and opportunities for further research.\\
%\textit{Method:} We carried out a systematic literature review with a view to finding the primary studies in the existing literature, which were later classified and analyzed according to four criteria: bibliometric key facts, research type facets, application scenario and knowledge graph characteristics. A subjective evaluation of the studies was also carried out to evaluate them in terms of methodology, empirical evidence, and relevance.\\
%\textit{Results:} As a result of the systematic literature review we found \textit{24} primary studies, published between October, 2016 and January, 2020. Most of them focus on ... They mainly consider top-down approaches and RDF knowledge graphs.\\
%\textit{Conclusions:} Existing research in the field is quite preliminary, and more research effort analyzing how knowledge graphs can be applied in the field of manufacturing and production is needed. Future research work should look at the possibility ...
%\end{abstract}

\begin{abstract}
Knowledge graphs in manufacturing and production aim to make production lines more efficient and flexible with higher quality output. This makes knowledge graphs attractive for companies to reach Industry 4.0 goals. However, existing research in the field is quite preliminary, and more research effort on analyzing how knowledge graphs can be applied in the field of manufacturing and production is needed. Therefore, we have conducted a systematic literature review as an attempt to characterize the state-of-the-art in this field, i.e., by identifying exiting research and by identifying gaps and opportunities for further research. To do that, we have focused on finding the primary studies in the existing literature, which were classified and analyzed according to four criteria: bibliometric key facts, research type facets, knowledge graph characteristics, and application scenarios. Besides, an evaluation of the primary studies has also been carried out to gain deeper insights in terms of methodology, empirical evidence, and relevance. As a result, we can offer a complete picture of the domain, which includes such interesting aspects as the fact that knowledge fusion is currently the main use case for knowledge graphs, that empirical research and industrial application are still missing to a large extent, that graph embeddings are not fully exploited, and that technical literature is fast-growing but seems to be still far from its peak.
\end{abstract}

\thispagestyle{empty}

\clearpage

%% file: 2_introduction.tex
\setcounter{page}{1}
%\twocolumn
%\linenumbers
\setlength{\parindent}{4em}
\setlength{\parskip}{1em}

\section{Introduction}
The twenty-first century has been clearly marked by its rapid growth in artificial intelligence (AI) applications. 
Thus, companies are required to undergo an inherent transformation to leverage AI for reaching Industry 4.0 standards and for gaining a competitive advantage on the international market. 
While AI technologies such as neural networks, natural language processing, chat-bots, autonomous driving vehicles, and digital twins received increasing attention in the field of manufacturing and production, little light is shed on the applications of knowledge graphs (KGs) in this domain.

% What is a knowledge graph
%% Open /academic KGs
In recent years, a large number of open (public) as well as closed (enterprise) KGs have been developed. 
While open KGs, which are often academic and open-source projects, provide access to anyone on the web, enterprise KGs are closed applications within companies that are only accessible to approved users~\cite{Blumauer_2020}.
Examples for public KG projects are DBpedia~\cite{lehmann2015}, Freebase~\cite{bollacker2008}, KBpedia~\cite{Bergman_2018},  NELL~\cite{carlson2010}, PROSPERA~\cite{nakashole2011}, Wikidata~\cite{vrandevcic2014}, and YAGO~\cite{suchanek2007}.
%% Closed /commercial KGs:
The most popular commercial and closed KGs are Cyc~\cite{lenat1995}, Google Knowledge Graph~\cite{singhal2012,Sullivan_2020}, Google Knowledge Vault~\cite{dong2014}, and Microsoft Satori. 
The blog entry by Google~\cite{singhal2012}\footnote{Note that Google's blog entry has been recently updated in~\cite{Sullivan_2020}.} is frequently quoted as seminal work of KG research since it sparked the discussion in this field in 2012. 
Considering early KG research from the 1980s, Google has rather revived KG technology than invented it. 
The foundation of KGs has been laid out by~\cite{sowa1983}, who provided conceptual graph theory as an early stage contribution for knowledge representation in semantic networks~\cite{sowa1992}. 
Further seminal work on KGs has been conducted by Stokman and co-authors, who aimed at building a KG to represent medical or sociological literature~\cite{bakker1987,smit1991,vries1989}.\footnote{The increasing popularity of knowledge graphs led to the funding of several projects related to knowledge graphs in manufacturing and other domains. 
See \url{https://www.nsf.gov/od/oia/convergence-accelerator/Award\%20Listings/track-a.jsp}.}
Even though knowledge graphs are nowadays frequently applied in different domains, there is still no formal definition, which is accepted in the entire community. In 2016, \cite{ehrlinger2016} proposed the following widely-acknowledged definition:  

[\textbf{Knowledge graph} by~\cite{ehrlinger2016}\label{def:kg}]
\textit{A knowledge graph acquires and integrates information into an ontology and applies a reasoner to derive new knowledge.}

% Definition~\ref{def:kg} has led to a broad discussion on ``what a knowledge graph actually is'', e.g., at the Dagstuhl Seminar 18371 on knowledge graphs~\cite{Bonatti_2019} as well as by~\cite{Fensel2020}. 
\cite{ehrlinger2016} further discuss existing alternative definitions and their implications and limitations. 
To be in-line with the state-of-the art, our paper covers all studies that claim to employ a knowledge graph if (1) they have been accepted by a scientific peer-reviewed journal or conference, or if (2) they conform to at least one definition reviewed by \cite{ehrlinger2016}. 
In summary, the common denominator of a KG is its structure in terms of nodes (entities) and edges (relationships). 
For storing graphs, the two most popular data models are \textit{RDF triples} and \textit{property graphs}\footnote{We refer to~\cite{Robinson_2015} for a discussion on less common graph data models, e.g., hypergraphs, which are not further considered in the frame of this paper.}. 
The majority of public KGs is stored in form of Resource Description Framework (RDF) triples (subject-predicate-object), proposed by the World Wide Web Consortium (W3C)\footnote{\url{https://www.w3.org/RDF}}. 
In RDF, subjects and objects are nodes and predicates the edges between the nodes. 
% A subject must be uniquely identified by an International Resource Identifier (IRI)
Property graphs store nodes and edges natively, whereas the nodes can have properties in form of key-value pairs~\cite{Robinson_2015}.

%% application domains for kgs
KGs are primarily used to semantically model a specific and often complex domain~\cite{Feilmayr_2016}. This explicitly modeled domain knowledge is used to support and enhance the accuracy of downstream tasks like question answering \cite{lukovnikov2017,zhang2018}, information extraction~\cite{daiber2013,dietz2018}, named entity disambiguation~\cite{zheng2012,zhu2018}, semantic parsing~\cite{berant2013,heck2013}, and recommender systems~\cite{sun2018,wang2019}. 
Also, the analysis of KGs with machine learning methods, e.g., to predict missing edges or to classify nodes, has gained increasing attention~\cite{Goyal_2018}. Since most machine learning models require a set of feature vectors as input, much research has been done to generate ``embeddings'' from KGs. A KG embedding transforms the nodes and (depending on the approach) also the edges to a numeric feature vector~\cite{Ristoski_2016}, which serves as direct input to a machine learning model.
Considering the plethora of application scenarios mentioned above, several domains have already perceived the substantial benefits KG technology brings with it. Example domains, which already rely on the use of KGs, are science~\cite{Auer_2018}, healthcare~\cite{ernst2014,abdelaziz2017,xie2018,li2020}, cybersecurity~\cite{iannacone2015,asamoah2016,han2018,deng2019}, data defects~\cite{Josko_2019}, education and training~\cite{sette2017,chi2018,dang2019}, and tourism~\cite{Fensel2020}.

% why is it novel/relevant topic?
This study aims to shine light on the state-of-the-art of KGs by revealing their utilisation in manufacturing and production. Already available standards and consensus with respect to the structure and construction of KGs in this domain should be discovered and discussed. 
According to the OECD\footnote{See, \href{https://data.oecd.org/trade/trade-in-goods-and-services.htm}{OECD}. G7 countries trade 2.323,096 million USD in services and 5.844,610 million USD in goods.}, more than 70~\% of the G7's world trade is based upon goods. 
Even though less than 25~\% of all jobs are provided by industries\footnote{See,~\href{https://data.worldbank.org/indicator/SL.IND.EMPL.ZS}{WorldBank}.}, a significant amount of jobs in other sectors depend on the jobs in the production sector.
Despite the substantial size and importance of the this sector, KGs have been neglected as one of the key AI technologies so far.
Thus, this research contributes to the dissemination and use of KGs in industry application by highlighting their benefits and how companies can leverage them. We aim at answering the research question: ``Which role play knowledge graphs in manufacturing and production?'' The question is answered by an investigation of (1) the bibliometric key facts, (2) research type facets, (3) KG characteristics, and (4) KG application scenarios.

The remainder of this paper is structured as follows: Section~\ref{literature} presents related work and Section~\ref{method} describes the planning and realization of the systematic literature review. 
The results, which were obtained from analyzing the primary studies and answers to the research questions are provided in Section~\ref{results}.
In Section~\ref{discussion}, we further discuss the research questions along with open research challenges and the threats to validity. Section~\ref{conclusion} concludes our study.

%% file: 3_related_work.tex
\section{Related Work} \label{literature}
% Does a similar mapping study or structured literature review exist?
To the best of our knowledge, there is no systematic literature review or systematic mapping study dedicated to knowledge graphs in manufacturing and production.
Yet, there are still surveys, reviews, and books that aim to provide an overview on the state-of-the-art of KG technologies.

Chronologically, we start with~\cite{nickel2015}, who provide the first survey on KGs with a special focus on the usage of latent and graph feature models for retrieving knowledge to predict new facts/edges in the graph. 
The foundation, architecture, construction, and applications of enterprise knowledge graphs is outlined in detail by~\cite{pan2017}.
\cite{paulheim2017} describes how to refine a knowledge graph based upon its A-box via completion, error detection, types of refinement, internal and external methods, and puts forward various evaluation standards that can be employed. 
The study of~\cite{wang2017} is similar to~\cite{nickel2015} with a comprehensive summary of translational distance and semantic matching models in the field of KG embeddings and a comment on the usefulness of KGs with respect to recommender systems and question answering applications. 
\cite{lin2018} employ a subset of the KG embeddings presented in~\cite{wang2017,nickel2015} and address complex relation modeling, relational path modeling, and multi-source information learning. 
Contrary to all previously mentioned articles, \cite{yan2018} does not focus on a specific KG topic, but gives a general overview on how KGs are constructed.\footnote{Keep in mind that this paper was submitted already in June, 2015 and got accepted for publication in January, 2016 which in turn means that it has been available prior \cite{nickel2015} which could explain the more general view on the topic matters.}
\cite{gesese2019} is the first survey of KG embeddings, which makes use of literals. 
Furthermore, \cite{kazemi2020} outlined how representation learning approaches are expedient for dynamic graphs. 
The book of \cite{kejriwal2019} demonstrates a very general summary on domain-specific KG construction. 
One survey about fault domain knowledge graphs has been written by \cite{wang2019a}. Another important scrutiny has been conducted by \cite{al2020} outlining preprocessing tools a la natural language processing, such as, named entity recognition, named entity disambiguation, and named entity linking, to enable the construction of a KG. \cite{Fensel2020} supply a very recent introduction into knowledge graphs with a lot of well-relatable real-life examples. 
\cite{heist2020} give an overview of cross-domain KGs that are publicly available on the Web. \cite{ji2020} extends the study of~\cite{wang2017} by explaining how different kind of neural networks can be used to generate KG embeddings. Finally, a  recent study by \cite{hogan2020} comprises all of the aforementioned studies' topics and provides a profound and comprehensive foundation into the field of knowledge graphs starting from scratch covering both, deductive and inductive knowledge representation techniques.

%% file: 4_method.tex
\section{Research Method} \label{method}

According to~\cite{brereton2007}, systematic literature reviews (SLRs) are ``\textit{a means of evaluating and interpreting all available research relevant to a particular research question or topic area or phenomenon of interest}''. SLRs are secondary  empirical  studies  used to  provide  a  structured  overview  of  a  research  field \cite{kitchenham2007guidelines}. An SLR follows a well defined methodology, which makes it less likely that the results of the literature are biased, although it does not protect against publication bias in the primary studies \cite{kitchenham2007guidelines}. 

In this systematic literature review, we followed the steps outlined in \cite{kitchenham2007guidelines} -- a systematic literature review consists of three main phases, i.e., the planning of the SLR, conducting the SLR, and reporting the SLR.  This section presents the planning of the study, i.e., the research questions, the data sources and search strategy, along with the classification and evaluation criteria.

\subsection{Research Questions}

The aim of this SLR is to analyze the current status of knowledge graphs in the field of manufacturing and production. 
Thus, existing research is investigated to identify potential gaps and opportunities for future work. The main research question guiding this study is:

\begin{center}
\textit{Which role play knowledge graphs in manufacturing and production?}
\end{center}

The research question we established for this study attempts to provide specific insights into the relevant aspects of how KGs are used in production and manufacturing. 
This includes questions about the articles' bibliometric key facts, research type facets, specific KG characteristics, and their application scenarios.
We also want to examine the type of research carried out up to that time (theoretical, proposal, empirical), together with the type of research forums in which these works have been published and presented. The exact research questions this systematic literature review answers are reported in Table~\ref{TableResearchQuestions}.

\begin{table}[!htbp] \centering 
\caption{Research questions} 
\label{TableResearchQuestions}
\begin{tabular}{@{\extracolsep{5pt}} ll} 
\\[-1.8ex]\hline
\bfseries Nr. & \bfseries Research questions \\
\hline \\[-1.8ex] 
RQ1 & What are the bibliometric key facts of KG publications? \\ 
RQ2 & Which research type facets do the identified publications address? \\
RQ3 & What are the specific knowledge graph characteristics?\\ 
RQ4 & What are application scenarios of knowledge graphs? \\ 
\hline \\[-1.8ex] 
%\multicolumn{3}{l}{Notes: } \\ 
\end{tabular} 
\end{table} 

RQ1 provides an overview of bibliometrics of published studies, concerning knowledge graph applications in production and manufacturing to exhibit the importance and timeliness of this topic. In more details, we analyze the publication trend, publication venues, and origin countries of research institutes that have published studies in this field. RQ2 investigates the maturity of knowledge graph applications by analyzing which research methods have been used for the validation of research. The specific construction techniques of knowledge graphs are addressed in RQ3. This is of major importance for consultants and practitioners as it reveals the structure of knowledge graphs employed in a production and manufacturing setting. Finally, RQ4 examines the application scenarios in which knowledge graphs have been used in the context of production and manufacturing, i.e., in which particular manufacturing domains knowledge graphs have been used, for which concrete use cases knowledge graphs are used, and which kinds of systems are developed based on knowledge graphs. 

\subsection{Data Sources and Search Strategy}
To build an adequate search string we have selected two major search terms:  \textit{‘Method’} and \textit{‘Field'}. The first major search term represents the employed methodology, namely, 'knowledge graph' whereas the second major search term illustrates the fields in which the knowledge graph should have been utilised. This term includes all sorts of technologies and synonyms of manufacturing and production in which the knowledge graph application should take place. Terms like 'enterprise', 'industry', 'company', 'corporate', 'manufacturer', 'manufacturing', 'organization', and 'production' should cover all synonyms for production and manufacturing, whereby, we included the German word 'industrie', as well, since knowledge graphs could also be applied in terms of \textit{Industrie 4.0} which is also commonly used in the international academic literature. Furthermore, 'internet of things' should supply us with references concerning 'internet of things' and 'industrial internet of things' whereas 'physical system' is related to literature with focus on the combination of knowledge graphs and cyber-physical systems. In addition, 'enterprise' and 'management' retrieve references with respect to 'enterprise knowledge graphs' and finally, we specifically outlined 'product' for 'product knowledge graphs' which is a rather new but interesting field of knowledge graph applications.

The final search string that has been used in the presented study is shown in Table \ref{TableSearchString}. The search terms were constructed using steps described in \cite{brereton2007}, in which the Boolean \textit{OR} is used to incorporate alternative spellings, synonyms or related terms, and the Boolean \textit{AND} is combining the link to major terms.

\begin{table}[!htbp] \centering 
  \caption{Search string} 
  \label{TableSearchString} 
\begin{tabular}{@{\extracolsep{5pt}} ll} 
\\[-1.8ex]\hline \\[-1.8ex] 
 & \bfseries Search terms\\
\hline \\[-1.8ex] 
Method & (``knowledge graph'') \\ 
& AND \\ 
Field & ((``enterprise'') OR \\ 
 & (``industry'') OR \\ 
 & (``industrie'') OR \\ 
 & (``physical system'') OR \\ 
 & (``internet of things'') OR \\ 
 & (``company'') OR \\ 
 & (``corporate'') OR \\ 
 & (``organization'') OR \\ 
 & (``product'') OR \\ 
 & (``production'') OR \\ 
 & (``management'') OR \\ 
 & (``manufacturer'') OR \\ 
 & (``manufacturing'')) \\ 
\hline \\[-1.8ex] 
%\multicolumn{3}{l}{Notes: } \\ 
\end{tabular} 
\end{table} 

The proposed search strategy is set out in Table \ref{TableSearchStragety}. The scope of the search considers publications and contributions presented in both academic and professional forums and publications. That is, we have considered academic publications (such as those published in journals or presented in academic conferences or peer-reviewed books) in addition to publications and contributions presented in industry or professional forums, such as conferences, workshops, and online publications. For academic publications, the sources of choice are: \textit{ACM Digital Library}, \textit{IEEExplore}, \textit{ISI Web of Science}, \textit{ScienceDirect} and \textit{Springer}. It has been a need to use a general search engine which in our case is \textit{Google Scholar} to include non-academic contributions and publications. Certain criteria on the data sources has been invoked to overcome particular challenges to avoid assessing hundreds of thousands of articles. To keep the search within reasonable bounds, we restricted the number of results retrieved from Google Scholar to 300\footnote{Note, that this number was sufficiently high, since a significant part of the last results returned by the engine did not include any primary studies.}. What is more, this data source was applied only to search for non-academic primary studies: those papers or articles published in industry/professional conferences, workshops, online journals/magazines or corporate blogs. The strategic search has been conducted recursively, that is, relevant studies referenced in the primary studies will also be considered. Personal blogs or web pages have been excluded from the search.

\begin{table}[!htbp] \centering 
  \caption{Summary of the search strategy} 
  \label{TableSearchStragety} 
  \small
\begin{tabular}{@{\extracolsep{5pt}} ll} 
\\[-1.8ex]\hline \\[-1.8ex] 
\bfseries Search strategy & \\
\hline \\[-1.8ex]
Academic databasese & ACM Digital Library\\ 
& IEEExplore\\ 
& ISI Web of Science\\ 
& ScienceDirect\\ 
& Springer\\ 
Other data sources & Google Scholar\\ 
Target items & Books\\ 
& Conference papers\\ 
& Workshop papers\\ 
& Journal papers\\ 
& Industry/professional workshop contribution\\ 
& Industry/professional conference contribution\\ 
& Non-academic online publications\\
Search applied to & Title\\
& Abstract\\
& Keywords\\
& Full-text (Google Scholar)\\
\hline \\[-1.8ex] 
%\multicolumn{3}{l}{Notes: } \\ 
\end{tabular} 
\end{table} 

The inclusion and exclusion criteria whether a paper is taken into account for the systematic literature review is shown in Table \ref{TableIECriteria}. Every study needs to include at least one of both major search terms. Additionally, it has to be published in an academic or professional forum. English has to be the language of the full-text and the publication date is not allowed to exceed the 26th of February, 2020. In case, the inclusion criteria have been fulfilled and none of the exclusion criteria has been triggered as well, the study will be considered as primary study in the systematic literature review. 

In the first round, the title and the abstract of each study is investigated whether the paper seems to be an eligible fit for the systematic literature review. Although corporate blog posts are considered, personal blogs or web pages are strictly excluded. In case, the paper is only available in the form of a \textit{PowerPoint} presentation or the emphasis of the article is not on knowledge graphs in a production or manufacturing setting it is excluded as well.

In the second round, we are left with all papers that have been affirmed to be relevant in the first round. In this round, also the full texts of the papers are considered. If an article only has an abstract but no full-text, or represents a summary of a workshop it is rejected. Non-academic or non-professional papers are eliminated as well. 
Further, we have dropped papers discussing knowledge graphs but only refer to manufacturing and production as potential usage domain without describing a concrete application scenario of knowledge graphs in this domain. For deciding if a paper describes a use case from the manufacturing domain we have used the \href{https://www.census.gov/eos/www/naics/}{North American Industry Classification System} (NAICS)\footnote{According to NAICS, the manufacturing sectors are, food manufacturing, beverage and tobacco product manufacturing, textile mills, textile product mills, apparel manufacturing, leather and allied product manufacturing, wood product manufacturing, paper manufacturing, printing and related support activities, petroleum and coal products manufacturing, chemical manufacturing, plastics and rubber products manufacturing, nonmetallic mineral product manufacturing, primary metal manufacturing, fabricated metal product manufacturing, machinery manufacturing, computer and electronic product manufacturing, electrical equipment, appliance, and component manufacturing, transportation equipment manufacturing, furniture and related product manufacturing, and miscellaneous manufacturing.} as a reference. The NAICS is a classification system for business establishments providing a systematic overview on the manufacturing domain. All other articles that are in-line with the inclusion and exclusion criteria are considered as primary studies.

\begin{table}[!htbp] \centering 
  \caption{Summary of the selection strategy} 
  \label{TableIECriteria} 
  \small
\begin{tabular}{@{\extracolsep{5pt}} ll} 
\\[-1.8ex]\hline \\[-1.8ex] 
\bfseries Inclusion/exclusion criteria & \\
\hline \\[-1.8ex]
Inclusion criteria & IC-1: Terms fulfill the search string\\ 
& IC-2: Academic journal, conference and workshop papers\\ 
& IC-3:  Contribution to conferences, workshops, and online publications\\
& IC-4: Papers written in English\\ 
& IC-5: Publication date: until 26th of February, 2020\\ 
Exclusion criteria & EC-1.1: Personal blogs or web pages\\
\quad for titles and abstract  & EC-1.2: Papers available only in the form of presentations\\
& EC-1.3: Papers which do not focus on knowledge graphs in manufacturing or production\\ 
\quad for full text & EC-2.1: Papers available only in the form of abstracts\\ 
& EC-2.2: Papers presenting a summary of a workshop\\
& EC-2.3: Non-academic/non-professional online publications\\
& EC-2.4: Papers which are using applications just as examples\\
& EC-2.5: Papers which do not focus on KGs in manufacturing according to the NAICS\\ 
\hline \\[-1.8ex] 
%\multicolumn{3}{l}{Notes: } \\ 
\end{tabular} 
\end{table} 

\subsection{Data Extraction and Synthesis}
To answer the RQs defined in Table \ref{TableResearchQuestions}, we extract specific data from the selected primary studies. Table \ref{TableExtractedInformation} highlights the data items ($D1$ to $D9$) extracted for the analysis in this review. $D1$, $D2$, and $D3$ provide clues concerning the distribution of knowledge graph studies in manufacturing and production over years, venues, and countries of publication, and thus answers $RQ1$. $D4$ and $D5$ directly contribute to the answers of $RQ2$. $D6$ can be used to answer $RQ3$. $D7$, $D8$, and $D9$ contribute to the answer of $RQ4$ and further discussion of knowledge graph approaches in manufacturing and production. To ensure that the data extraction results are unbiased, two authors performed the data extraction for each primary study independently, and then one checked the data extraction results of the other, and finally they discussed and reached a consensus on the data extraction results.

\begin{table}[!htbp] \centering 
  \caption{Data items extracted from each study} 
  \label{TableExtractedInformation}
  \small
\begin{tabular}{@{\extracolsep{5pt}} llll} 
\\\hline \\[-1.8ex] 
\bfseries Nr. & \bfseries Item name & \bfseries Description & \bfseries Relevant RQ \\
\hline \\[-1.8ex] 
D1 & Publication year & In which year was the article published? & RQ1\\
D2 & Venue & What is the name of the journal, conference, or workshop? & RQ1 \\
D3 & Country & Where are the research institutes located that have published the study? & RQ1 \\
D4 & Research field & What are the \href{https://www.scimagojr.com/journalrank.php}{Scimago} classifications of the outlet? & RQ2 \\
D5 & Evidence level & What is the evidence level of the evaluation of the proposed approach? & RQ2\\
D6 & KG approach & What kind of knowledge graph and which technique has been used? & RQ3\\
D7 & Domain & In which manufacturing domain has the knowledge graph been constructed? & RQ4 \\
D8 & Use Case & Which use case is supported by the knowledge graph? & RQ4 \\
D9 & System Kind & Which kind of system has been developed based on a knowledge graph? & RQ4 \\
\hline \\[-1.8ex] 
%\multicolumn{3}{l}{Notes: } \\ 
\end{tabular} 
\end{table} 

\subsection{Evaluation}
A six-point Likert-scale was designed to provide a quality assessment of the selected primary studies. We categorize studies to five different research type facets that are weighted according to their quality of evidence (as proposed by \cite{wieringa2006requirements}), namely from \textit{Opinion Papers} '1' to \textit{Evaluation Research} '5'. The final numerical value which generates the evaluation of each paper assumes a value between 0 and 1. The evaluation provides insights into the degree to which different aspects of knowledge graphs are considered in existing research in the field. It was decided that the results of this assessment would help to identify the quality of research carried out.

The questions composing the quality assessment are shown in Table~\ref{TableEvaluationQuestions} and follow the six-level classification of evidence evaluation suggested by \cite{alves2010}. The purpose of these evaluation questions was to assess the primary studies based upon the employed methodology, as well as, how the proposal has been integrated. 

\begin{table}[!htbp] \centering 
  \caption{Evaluation Level of Research Type Facet (see \cite{wieringa2006requirements})}
  \label{TableEvaluationQuestions} 
\begin{tabular}{lp{1.2cm}p{12cm}} 
\\
\hline\\[-1.8ex]
\bfseries Research Type Facet & \bfseries Evidence Level & \bfseries Description \\
\hline \\[-1.8ex]
Evaluation Research & 1.0 & Techniques are implemented in practice and an evaluation of the technique is conducted. That means, it is shown how the technique is implemented in practice (solution implementation) and what are the consequences of the implementation in terms of benefits and drawbacks (implementation evaluation). This also includes to identify problems in industry. \\ 
Validation Research & 0.8 & Techniques investigated are novel and have not yet been implemented in practice. Techniques used are for example experiments. \\
Solution Proposal & 0.6 & A solution for a problem is proposed, the solution can be either novel or a significant extension of an existing technique. The potential benefits and the applicability of the solution is shown by a small example or a good line of argumentation.\\ 
Philosophical Papers & 0.4 & These papers sketch a new way of looking at existing things by structuring the field in form of a taxonomy or conceptual framework.\\ 
Experience Papers & 0.2 & Experience papers explain on what and how something has been done in practice. It has to be the personal experience of the author. \\
Opinion Papers & 0.0 & These papers express the personal opinion of  somebody whether a certain technique is good or bad, or how things should be done. They do not rely on related work and research methodologies. \\ 
\hline \\[-1.8ex] 
%\multicolumn{3}{l}{Notes: } \\ 
\end{tabular} 
\end{table}

%% file: 5_results.tex
\section{Empirical Results} \label{results}
In this section, we offer the detailed results of our literature analysis.  Thus, this section is structured around the four research questions we have answered. 

\subsection{Results of the Search}
The strategic search resulted in \textit{833} articles. Detailed break down by databases is shown in Table \ref{TableSearchResults}. We have obtained \textit{227}, \textit{172}, \textit{84}, \textit{23}, \textit{21}, and \textit{356} studies from \textit{ACM Digital Library}, \textit{IEEExplore}, \textit{ISI Web of Science}, \textit{ScienceDirect}, \textit{Springer}, and \textit{Google Scholar}\footnote{In the case of Google Scholar, two searches have been conducted, whereas the first search targeted all studies that have the search string included in the title of the article (56 articles), and the second search extracted the first 300 most relevant articles that include the search string in the manuscript.}, respectively.

\begin{table}[!htbp]
  \centering 
  \caption{Electronic databases included in this Mapping Study} 
  \label{TableSearchResults} 
\begin{tabular}{@{\extracolsep{-5pt}} lll} 
\\[-1.8ex]\hline \\[-1.8ex] 
\bfseries Nr. & \bfseries Database & \bfseries Number of studies \\
\hline \\[-1.8ex]
DB1 & ACM Digital Library & 227\\ 
DB2 & IEEExplore & 172\\ 
DB3 & ISI Web of Science & 84\\ 
DB4 & ScienceDirect & 23\\ 
DB5 & Springer & 21\\ 
DB6 & Google Scholar & 356\\ 
\hline \\[-1.8ex] 
%\multicolumn{3}{l}{Notes: } \\ 
\end{tabular} 
\end{table}

\begin{figure}[!htbp]
\centering
\caption{Primary study selection process}
\label{FigSelection}
\includegraphics[width=0.8\maxwidth]{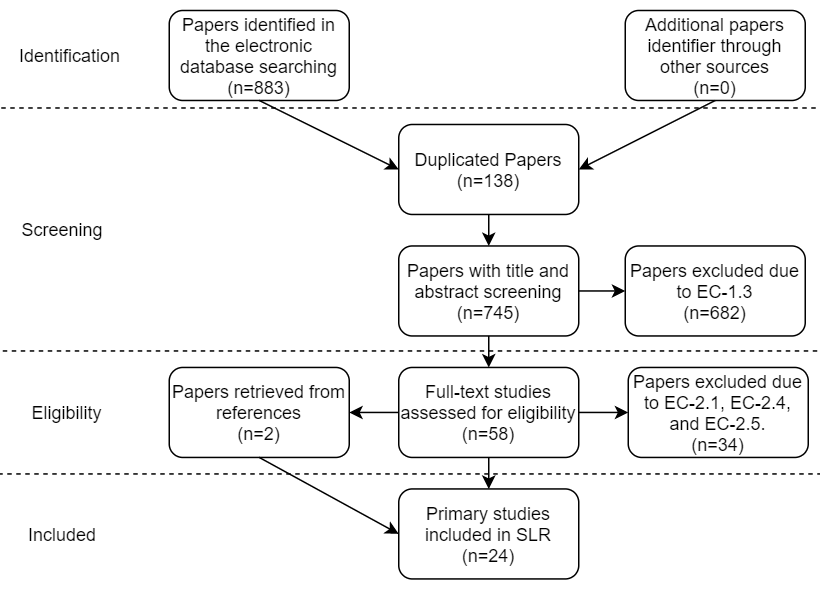}
\end{figure}

Removing the duplicates of all \textit{833} detected publications leaves us left with \textit{745} studies. Out of those \textit{745} studies, the abstract has revealed that \textit{682} are not relevant for the systematic literature review. \textit{5} out of the remaining \textit{63} articles have not been written in English. After reading the full-text of all \textit{58} articles, \textit{24} have been identified to be relevant as primary studies for this systematic literature review. In each step, at least two authors needed to have the same opinion on whether a paper gets included or excluded. In case, two authors had a different opinion on one study, the third author decided whether to keep the primary paper in the selection process. \textit{2} articles have been added as a result of snowballing the references of all papers. The detailed breakdown is given in Figure \ref{FigSelection}. As the findings indicate, the number of primary studies obtained may appear to be quite small – there are just \textit{24}, however, as will be shown in greater detail in this section, all these papers were published between the years 2016 and 2020. The full list of primary studies gathered is listed in the appendix.

\subsection{Research Questions}
In this section, we answer each research question outlined in Table~\ref{TableResearchQuestions} by discussing the analysis of the primary studies. 
% an analysis of the primary studies is performed that is in-line with the classification criteria and research questions outlined previously.

\subsubsection{RQ1: What are the bibliometric key facts of KG for publications?} \label{rq1}
\begin{figure}[!htbp]
\centering
\caption{Distribution of primary studies by year}
\label{FigYear}
\includegraphics[width=0.5\maxwidth]{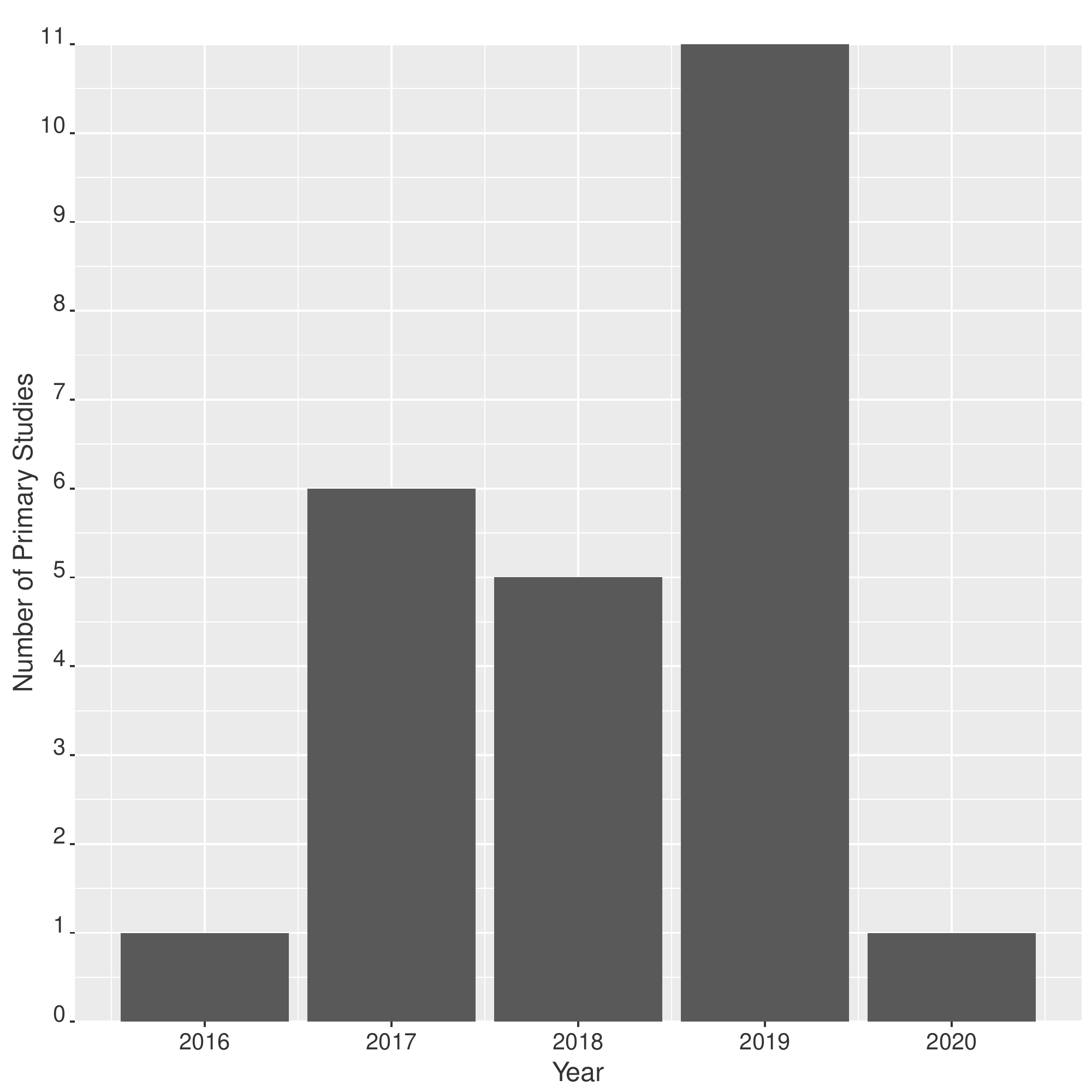}
\end{figure}

Figure \ref{FigYear} shows the distribution of the studies according to the year they have been published. The first primary studies with a focus on KGs in a manufacturing or production environment date back to 2016. This highlights the fact that this field is still in its early stages of development. The number of studies published in 2017 is six times that of 2016. In addition, the number of publications in 2017 is larger than in 2018. However, the number of primary studies published in 2019 is more than twice as much as in 2018. As we have only covered the first two months in 2020 the number of published studies can be disregarded, however, deducting from previous years we expect that KG in production and manufacturing studies will exceed 2019 as a result of its increasing popularity. 

\begin{figure}[!htbp]
\centering
\caption{Distribution of primary studies by country}
\label{FigCountry}
\includegraphics[width=0.5\maxwidth]{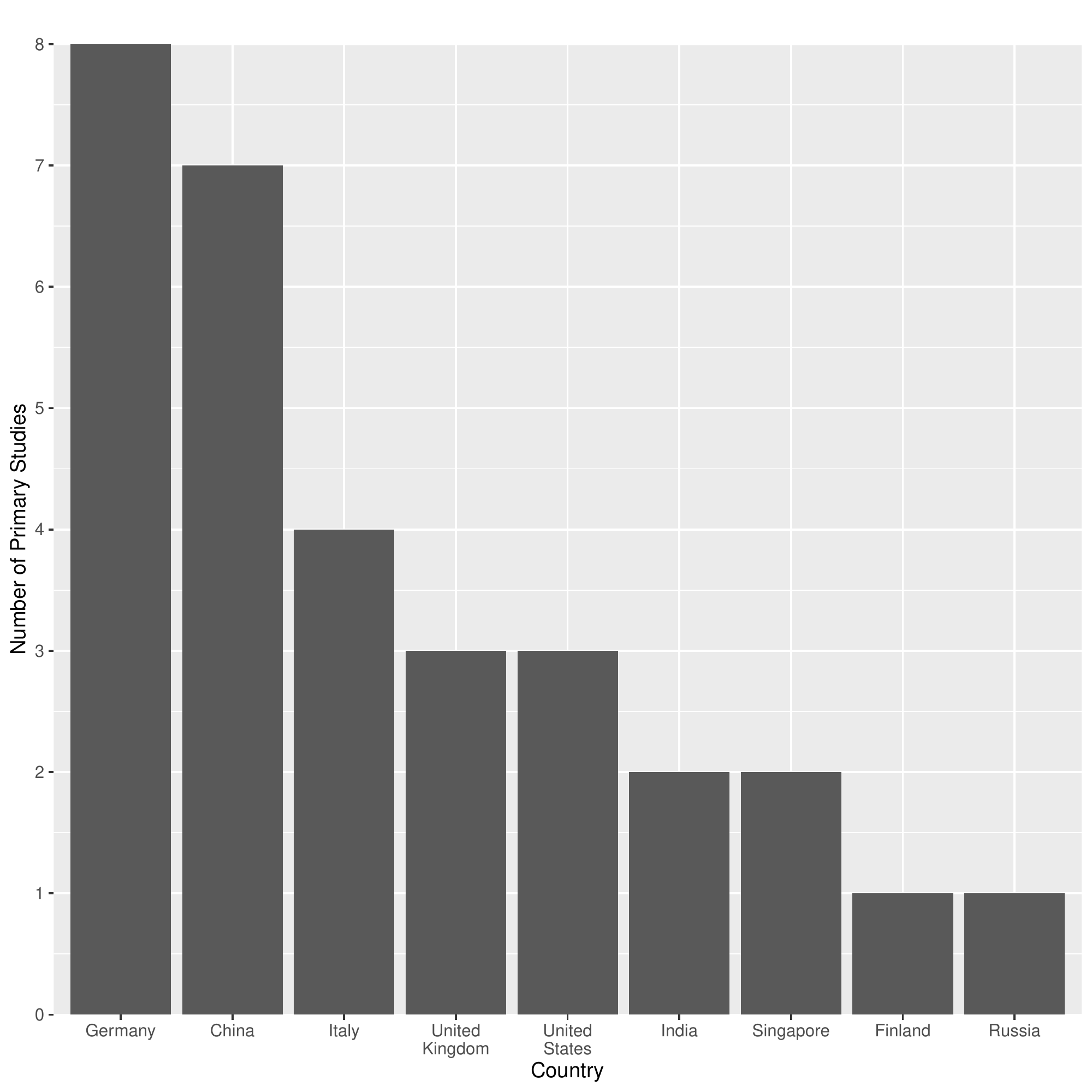}
\end{figure}

Next, we have analyzed the origin countries of the primary studies. Therefore we  have analyzed the affiliations of the authors, i.e, the countries of their research institutes. For each study, we have taken the origin countries of all authors into consideration. Figure~\ref{FigCountry} depicts the results of the analysis and shows that most primary studies originate from research institutes located in Germany (8 studies), and China (7 studies). Four studies were published by research institutes from Italy, followed by three studies from the United Kingdom and the United States each. Both, India and Singapore are the origin of two studies, and one study originates from Finland and Russia each. 

\begin{figure}[!htbp]
\centering
\caption{Distribution of primary studies by forum}
\label{FigForum}
\includegraphics[width=0.5\maxwidth]{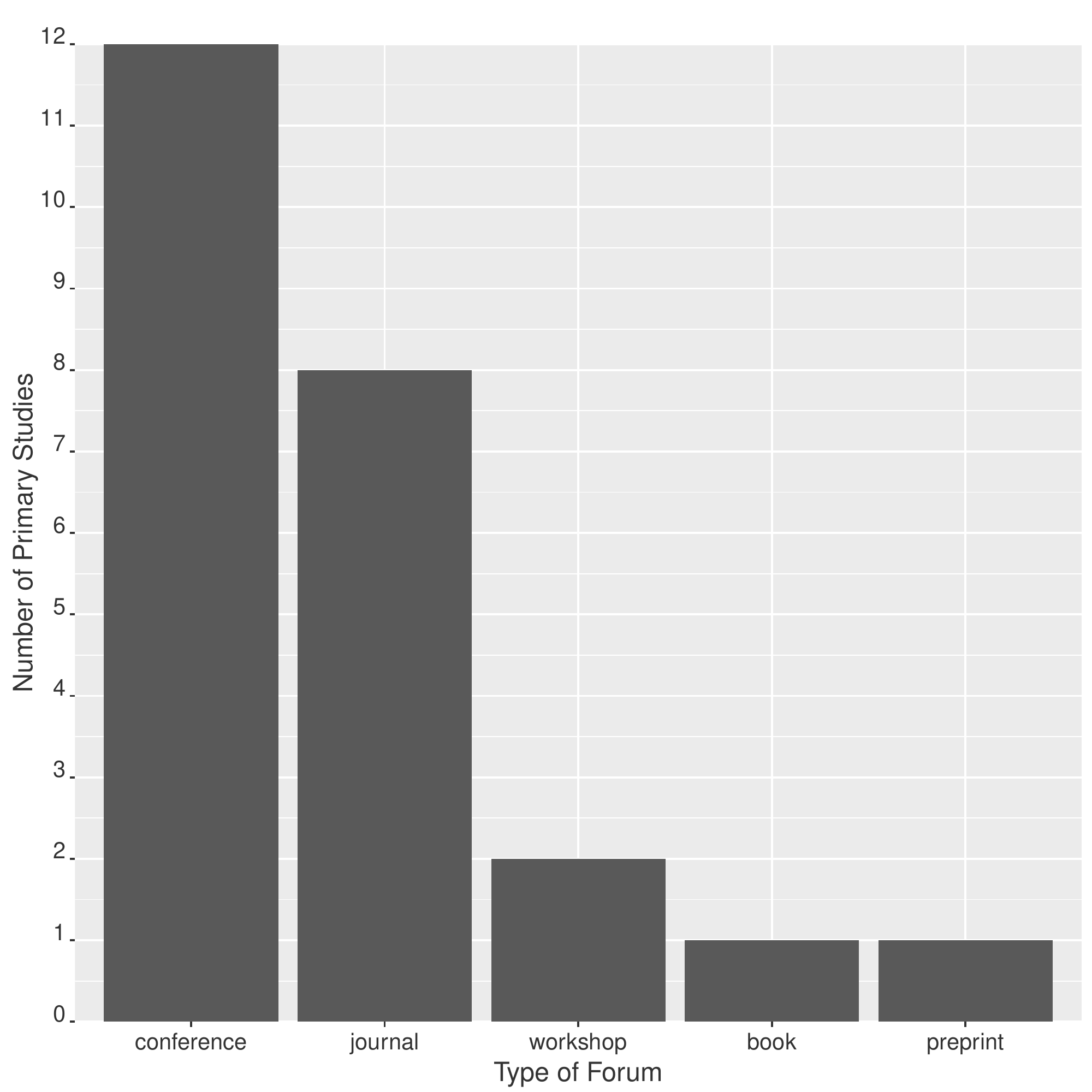}
\end{figure}

After the analysis of the primary studies by country, we continue by focusing on the type of publication forum. The ordered distribution of the type of forums is shown in Figure \ref{FigForum}. A closer look at Figure \ref{FigForum} points out that \textit{12} primary studies -- which are exactly half of all primary studies -- have been published in conference proceedings. \textit{8} primary studies -- representing one-third of all studies -- have found their outlet in journals. In addition, only \textit{2} articles have been printed in workshops whereas \textit{1} primary study has been available as a book, and \textit{1} primary study is still a pre-print.

\begin{figure}[!htbp]
\centering
\caption{ Distribution of primary studies by Scimago research field}
\label{FigField}
\includegraphics[width=0.95\maxwidth]{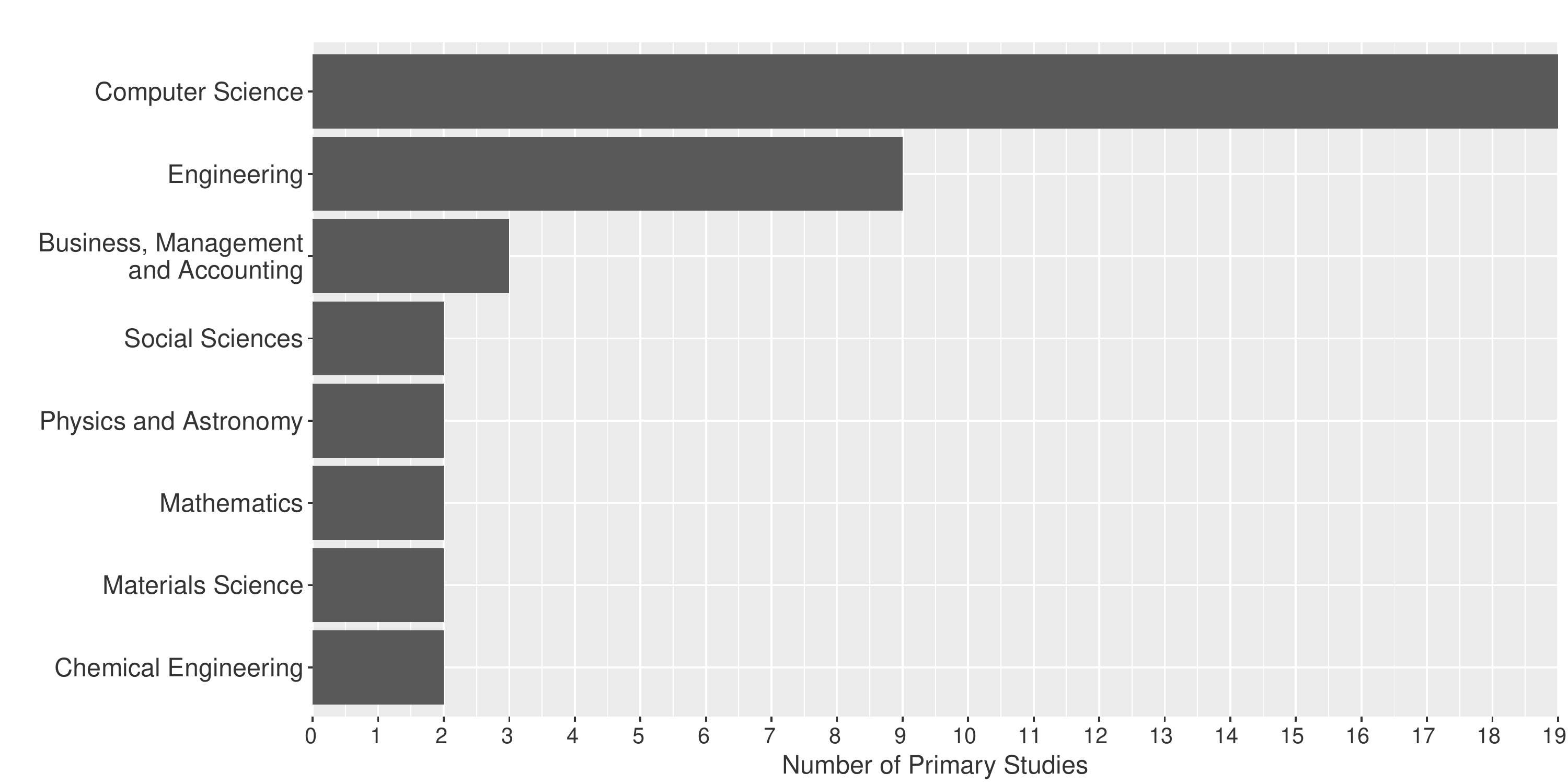}
\end{figure}

We continue by classifying the primary studies according to their Scimago\footnote{\url{https://www.scimagojr.com}} research field. The classification by Scimago is not unique and therefore, a single study can be assigned to multiple research fields. 
Figure~\ref{FigField} reveals that most studies have been published in \textit{Computer Science}. In more detail, \textit{19} articles and hence almost 80~\% of the articles belong to this category. \textit{9} primary studies are considered in the field of \textit{Engineering} which represent 37.5~\% of all selected papers. Additionally, \textit{3} studies have been published in outlets that are considered in the field of \textit{Business, Management and Accounting}. Finally, all remaining fields -- \textit{Chemical Engineering}, \textit{Materials Science}, \textit{Mathematics}, \textit{Physics}, and \textit{Astronomy}, and \textit{Social Science} -- inherent \textit{2} primary studies. This demonstrates that KGs mainly belong to the discipline of computer science and engineering.

\subsubsection{RQ2: Which research type facets do the identified publications address?} \label{rq2}
The question regarding the research type facet is of fundamental importance as it demonstrates the applicability of the approach and whether it has already been implemented in a real-world scenario. This information is not only essential for researchers, but it is also of major concern for practitioners and consultants.

\begin{figure}[!htbp]
\centering
\caption{Distribution of primary studies by research type facet}
\label{FigResearch}
\includegraphics[width=0.5\maxwidth]{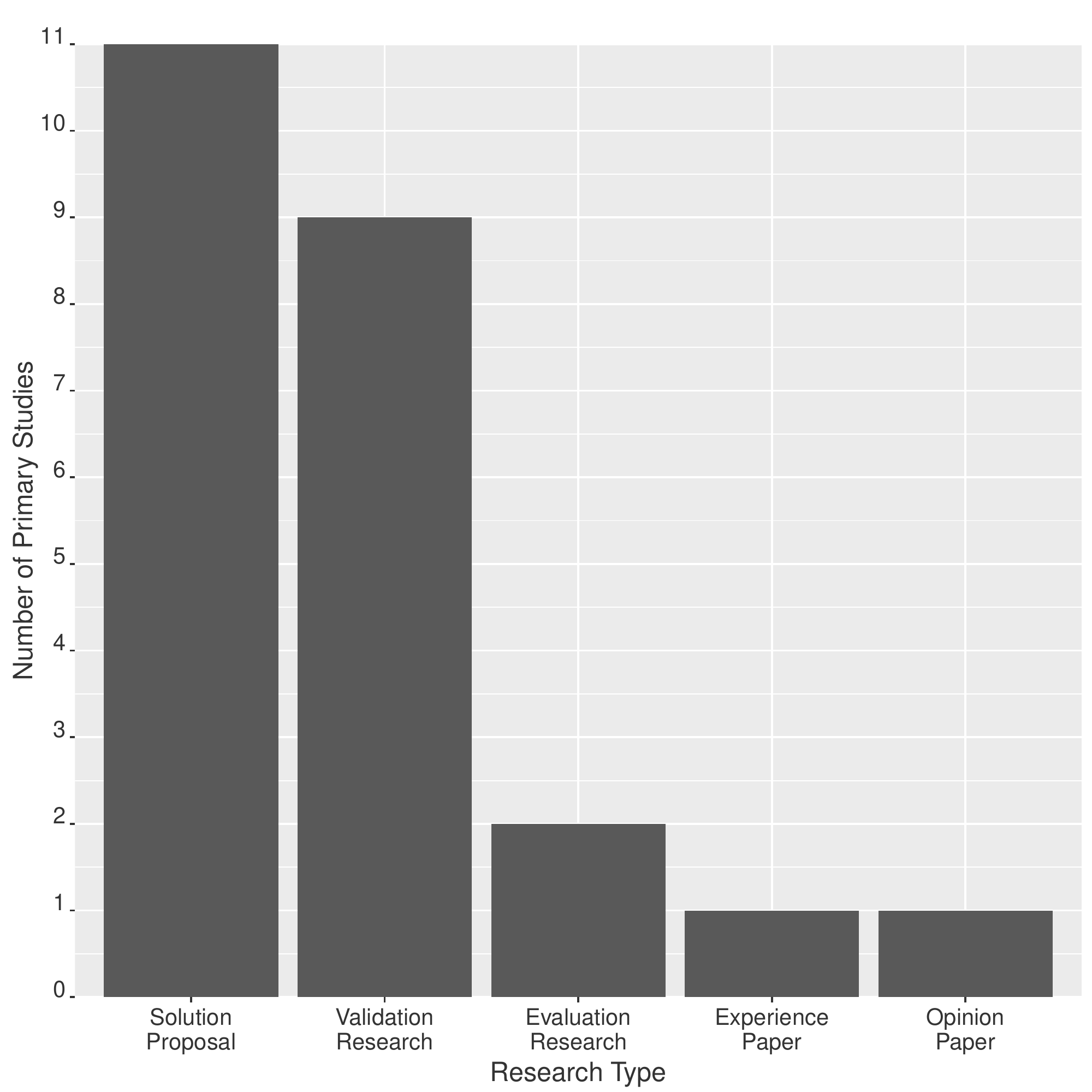}
\end{figure}

Figure \ref{FigResearch} explicitly displays the clear picture of the primary studies by research facet type. In this regard, the majority of primary studies -- \textit{11} in total -- can be considered as solution proposals since although they have paved the ground for novel KG applications, neither evaluation nor validation took place. \textit{9} selected articles fall into the category of validation research. Hereby, the knowledge graph methodology has been employed in a real-world setting. More information about the specific application scenario is given in Section \ref{rq4}. Less attention has been brought to evaluation research since only \textit{2} papers have been conducted so far. In both remaining categories -- experience papers and opinion papers -- we have found a single study. This analysis shows that even though multiple solutions for industry-specific problems have been proposed, there is not a wealth of literature yet that has incorporated those suggestions to real-world applications. This step is clearly missing at the current state of the literature. 

The detailed breakdown is illustrated in Table \ref{ResearchTypeFacetsTable}:

\begin{table}[!htbp]
\centering
\renewcommand{\arraystretch}{1.3} \caption{Classification of Primary Studies by Research Type Facet}
\label{ResearchTypeFacetsTable}
\centering
\begin{tabular}{@{\extracolsep{-5pt}} ll} 
\hline
\bfseries Research Type Facet & \bfseries Studies \\\hline
Evaluation Research & \citeS{5} \citeS{22} \\
Validation Research & \citeS{3} \citeS{4} \citeS{11} \citeS{12} \citeS{13} \citeS{14} \citeS{17}  \citeS{23} \citeS{24} \\
Solution Proposal & \citeS{1} \citeS{2} \citeS{6} \citeS{7} \citeS{9} \citeS{10} \citeS{15} \citeS{18} \citeS{19} \citeS{20} \citeS{21}\\
Experience Paper & \citeS{16} \\
Opinion Research & \citeS{8} \\\hline
\end{tabular}
\end{table}

\subsubsection{RQ3: What are the specific knowledge graph characteristics?} \label{rq3}
In this section, we examine the type and construction of the employed KGs. According to literature, a KG can either be constructed by a bottom-up or a top-down approach~\cite{kaufmann2014,zhao2018}. 
In the top-down approach, a domain expert conceptually models the schema of the KG, often in form of an ontology, which is then populated with data to complete the knowledge graph. In the bottom-up approach, the structure of the knowledge graph is (typically automatically) induced from the data~\cite{kaufmann2014,zhao2018}. 
In terms of data model, a KG can either be an RDF graph or a property graph~\cite{das2014,hartig2014,alocci2015,angles2019,Robinson_2015}. 

The decision for a specific construction approach and data model clearly influences a successful application of the KG. The top-down approach, for instance, is very restrictive when adding new triples due to carefully modeled schema constraints. The bottom-up approach, on the other hand, allows a rapid growth of the KG with little human effort, which could however lead to KGs with lower quality (e.g., incorrect relationships). Interestingly, we did not find any paper, which used machine learning or deep learning approaches to construct a KG.

\begin{figure}[!htbp]
\centering
\caption{Distribution of kind of knowledge graph}
\label{FigKind}
\includegraphics[width=0.5\maxwidth]{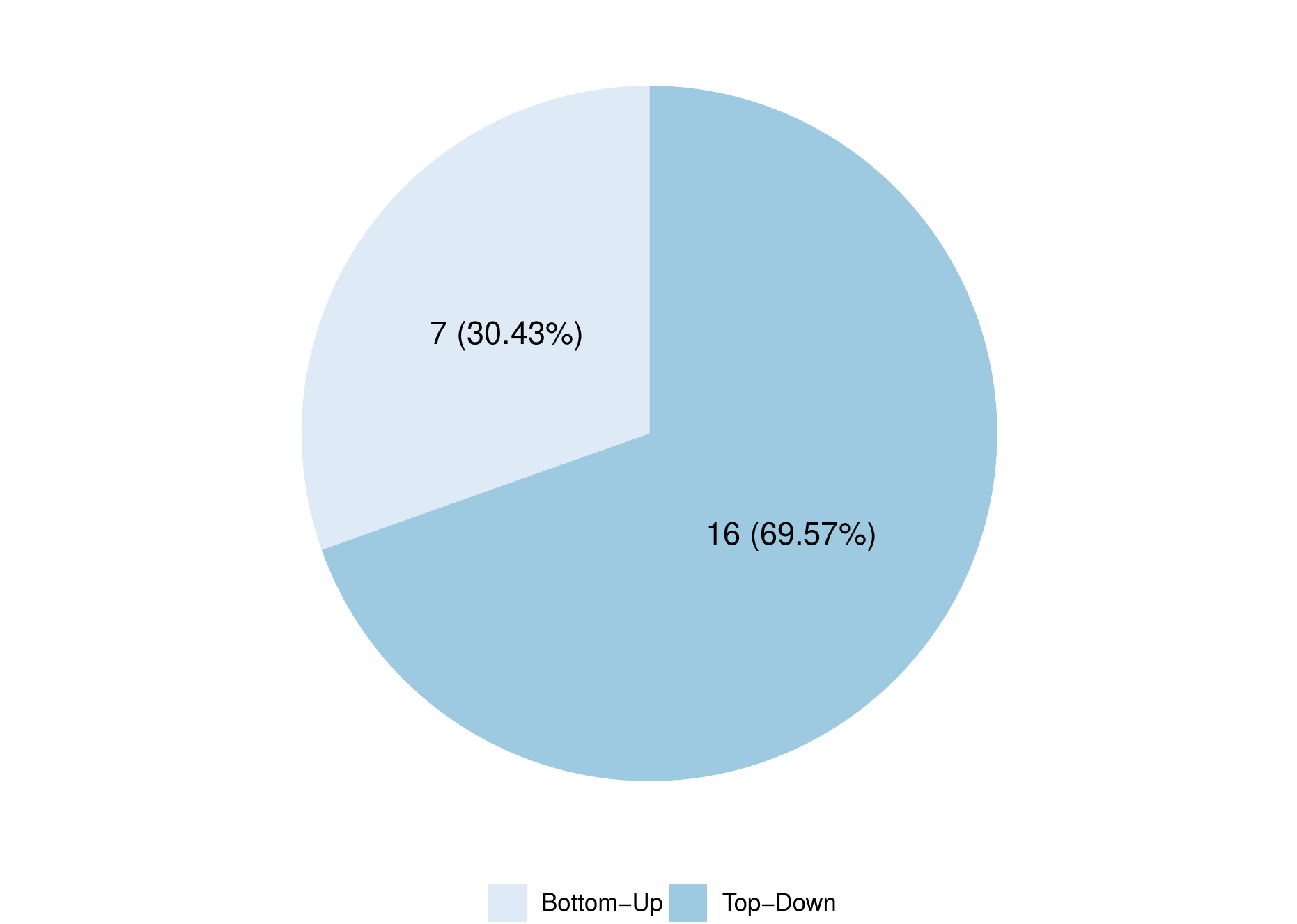}
\end{figure}

Figure~\ref{FigKind} shows that 69.57~\% of the KGs have been constructed with a top-down approach and only 30.43~\% with a bottom-up approach. 
This is an interesting finding, since a growing amount of research focuses on the automated generation of KGs with bottom-up approaches. For example, \cite{wang2017} use neural networks to induce the structure of a KG.

\begin{figure}[!htbp]
\centering
\caption{Distribution of knowledge graph construction}
\label{FigConstruction}
\includegraphics[width=0.5\maxwidth]{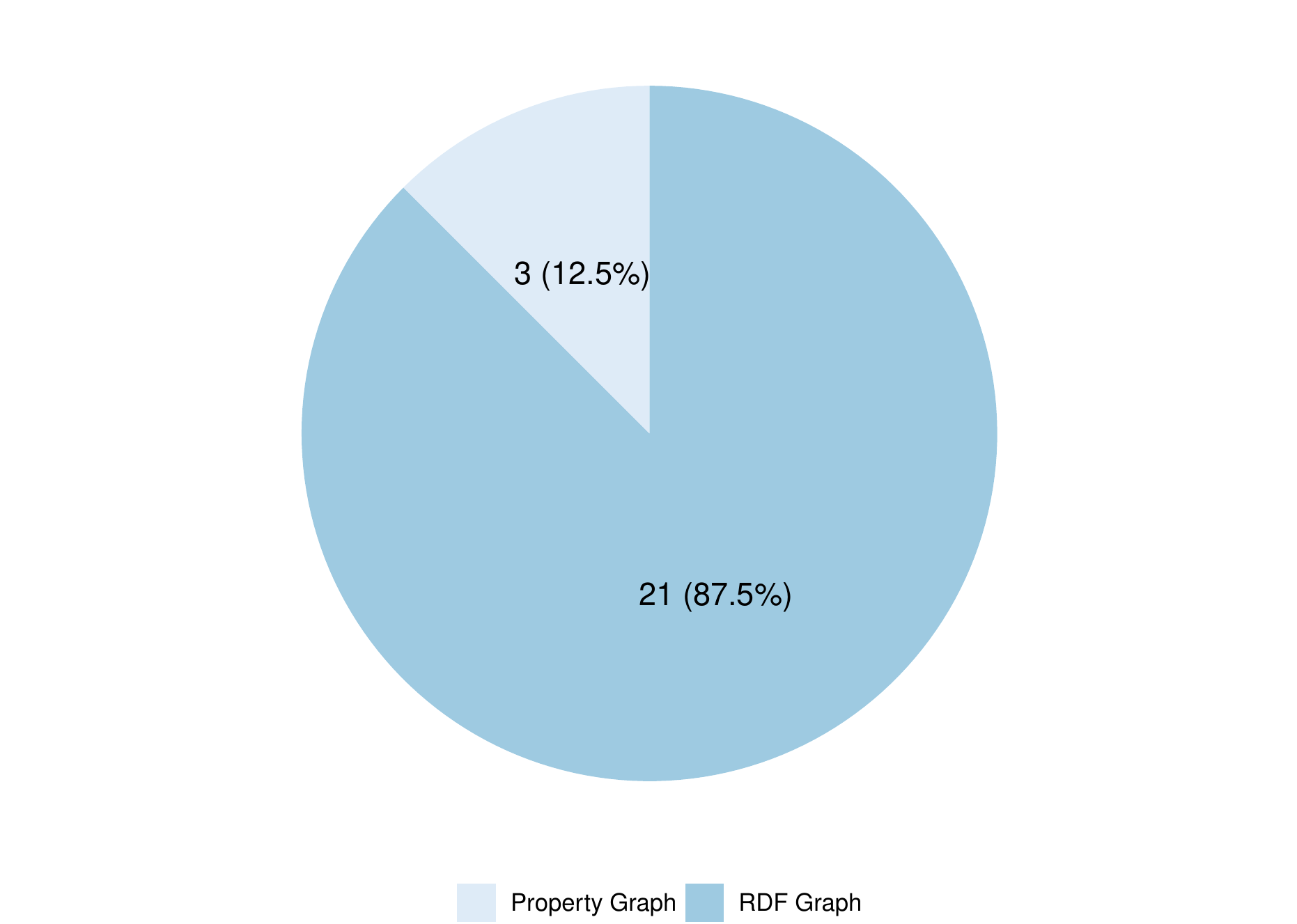}
\end{figure}

With respect to the data model, Figure~\ref{FigConstruction} discloses that the majority of KGs (87.5~\%) are modeled as RDF graphs and only 12.5~\% are modeled as a property graphs. Even though, 12.5~\% is not much, it is still more than expected since there is a strand of literature that claims that knowledge graphs are RDF graphs \cite{ehrlinger2016}.

%Furthermore, we found that \textit{11} out of {24} primary studies have employed a reasoning component. 

\subsubsection{RQ4: What are the application scenarios of knowledge graphs?}  \label{rq4}

We have analyzed the primary studies with regard to application domain, use case, and system kind. 
Table~\ref{TabSummaryTable} provides a summary of each primary study.

\begin{figure}[!htbp]
\centering
\caption{Knowledge Graph Manufacturing Domains}
\label{FigDomain}
\includegraphics[width=0.95\maxwidth]{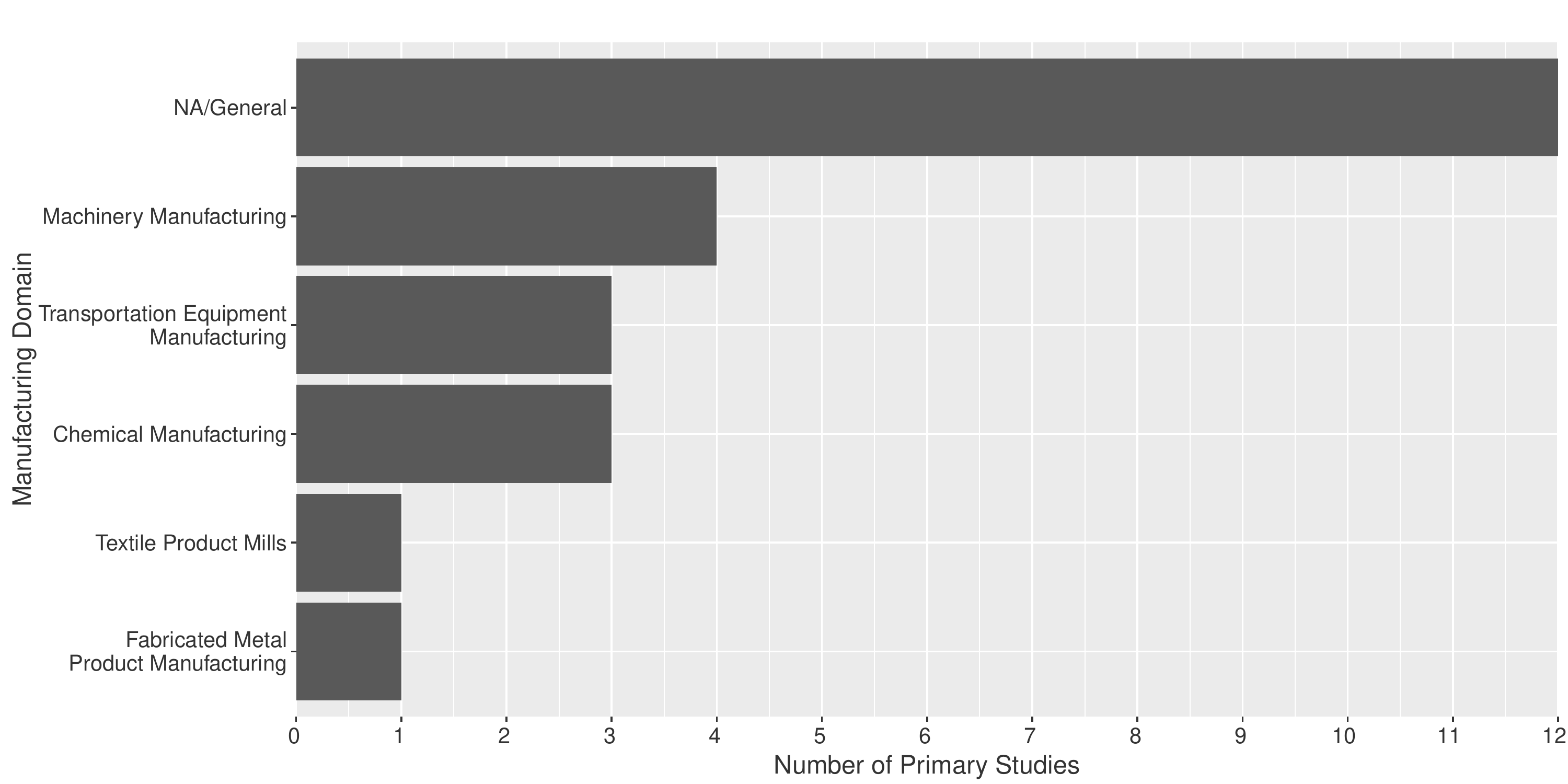}
\end{figure}

The application domain describes in which field of manufacturing and production KGs have been used. Figure \ref{FigDomain} depicts the results of the classification with respect to the NAICS schema. As shown in the figure, in half of the analyzed primary studies - 12 primary studies - no concrete application domain was provided. 4 primary studies describe the use of KGs in the field of \textit{machinery manufacturing}, 3 primary studies come from the field of \textit{chemical manufacturing}, 3 primary studies come from the field of \textit{transportation equipment manufacturing}, 1 primary study has been published in the field of \textit{fabricated metal product manufacturing}, and 1 primary study comes from the context of \textit{textile product mills}. The assignment of each primary study to an application domain is shown in Table \ref{DomainsTable}.

\begin{center}
\begin{longtable}{l p{.80\textwidth} }
\caption{Summary of Primary Studies} \\\hline
\label{TabSummaryTable}
\bfseries Primary Study & \bfseries Summary \\\hline
\citeS{5} & Development of a KG based on data of a fashion manufacturer. The KG integrates different knowledge sources, e.g., data of different products and sales data. On top of the KG a data visualization tool is developed supporting dynamic data analysis and sales forecasting.  \\
\citeS{22} & Construction of a KG based on big data in the field of additive manufacturing as foundation for a data analysis platform.  \\
\citeS{3} & Construction of a KG representing a digital twin of an automated production line. The KG integrates contextual data (e.g., equipment configuration) with operational data (process data and events, e.g., alarms). A machine learning component adapts the KG in response to changes of manufacturing environment. \\
\citeS{4} & Generation of a KG based on sensor data of a production line and inference of additional relationships based on an ontology.  \\
\citeS{11} & A framework for the construction of a KG based on descriptions of cyber-physical systems described using smart manufacturing standards, e.g., AutomationML.  \\
\citeS{12} & Construction of a KG representing a cyber-physical system based on multiple descriptions of different perspectives of the CPS, which are semantically integrated with each other. \\
\citeS{13} & Definition of a KG storing manufacturing knowledge and production problems. On top of the knowledge graph, a knowledge platform is constructed. The platform provides a knowledge matching component that links production problems to related manufacturing knowledge. \\
\citeS{14} & In this work a KG is used for the generation of action plans for autonomous robots. The KG is comprised of environment data, robot capabilities, task templates, and goal descriptions.  \\
\citeS{17} & A general purpose KG platform based on open standards. The platform has been evaluated in an industrial setting for integrating static and runtime data at a company from the manufacturing domain.  \\
\citeS{23} & Use of the J-Park Simulator for the construction of a distributed KG for the simulation of an air pollution scenario from the process industry.  \\ 
\citeS{24} & Creation of a KG from natural language requirement documents with the goal to support the automated generation of test cases from these documents. Application of the approach in the automotive domain. \\ 
\citeS{1} & Construction of a KG representing a digital twin of an automated production line. The KG integrates contextual data (e.g., equipment configuration) with operational data (process data and events, e.g., alarms). A machine learning component adapts the KG in response to changes of manufacturing environment. \\
\citeS{2} & Construction of an KG for metallic material. The KG is constructed based on DBpedia data, which is then enriched with data from wikipeda. Data of the KG can be displayed via a dedicated prototype.  \\
\citeS{6} & Construction of an open industrial KG by integrating multiple knowledge source (i.e., technical documents, data models). The KG supports the intelligent search for manufacturing services, by searching and evaluating manufacturers of machinery parts.  \\
\citeS{7} & Construction of an industrial KG as foundation for a digital twin based on multiple data sources, e.g., machine data, process data, material data. A machine learning component analyzes the KG and event data (log files) to complete the KG with additional links. \\
\citeS{9} & Development of a KG in the field of electrical discharge machining by integrating different data sources in a knowledge graph. Automated matching of identical concepts by applying intelligent matching algorithms that are used for the automated construction of an ontology.  \\
\citeS{10} & Development of a KG from CAM Models. The KG integrates the CAD models with NC process data of the CAM models and support the interchange of different CAM Models via ontology matching. \\
\citeS{15} & Use of the J-Park Simulator for the construction of a KG as foundation for the development of a smart contract-based market place. The system supports the automated search for services including service rating, service contracting, and service payment.   \\
\citeS{18} & Use of a KG as foundation of a single shared information space for describing objects of a manufacturing enterprise.  \\
\citeS{19} & Construction of a KG from unstructured Chinese text from multiple heterogeneous data sources for the automotive industry. The graph is build by applying natural language processing and deep learning for the identification of entities and relationships \\
\citeS{20} & Use of a KG for automated code generation in the field of automated guided vehicles.  \\ 
\citeS{21} & Construction of a KG from isolated data silos to support consistent data provision and to improve factory planning processes. \\
\citeS{16} & In this work a KG is used for the generation of action plans for autonomous robots. The KG is comprised of environment data, robot capabilities, task templates, and goal descriptions.  \\
\citeS{8} & Use of KG as foundation for machine learning based on graph embeddings. \\
\hline
\end{longtable}
\end{center}

\begin{table}[!htbp] 
\centering
\renewcommand{\arraystretch}{1.3} \caption{Manufacturing Domains of Primary Studies}
\label{DomainsTable}
\centering
\begin{tabular}{@{\extracolsep{-5pt}} ll} 
\hline
\bfseries Domain & \bfseries Studies \\\hline
NA / General &  \citeS{3} \citeS{4} \citeS{11} \citeS{12}  \citeS{13} \citeS{14}  \citeS{17} \citeS{6}  \citeS{7}   \citeS{18} \citeS{20} \citeS{8} \\
Machinery Manufacturing & \citeS{22} \citeS{9} \citeS{10} \citeS{16}    \\
Chemical Manufacturing & \citeS{23} \citeS{1} \citeS{15}    \\
Transportation Equipment Manufacturing &  \citeS{24} \citeS{19} \citeS{21}  \\
Fabricated Metal Product Manufacturing & \citeS{2} \\
Textile Product Mills & \citeS{5}  \\\hline
\end{tabular}
\end{table}

\begin{figure}[!htbp]
\centering
\caption{Knowledge Graph Use Cases}
\label{FigUseCase}
\includegraphics[width=0.95\maxwidth]{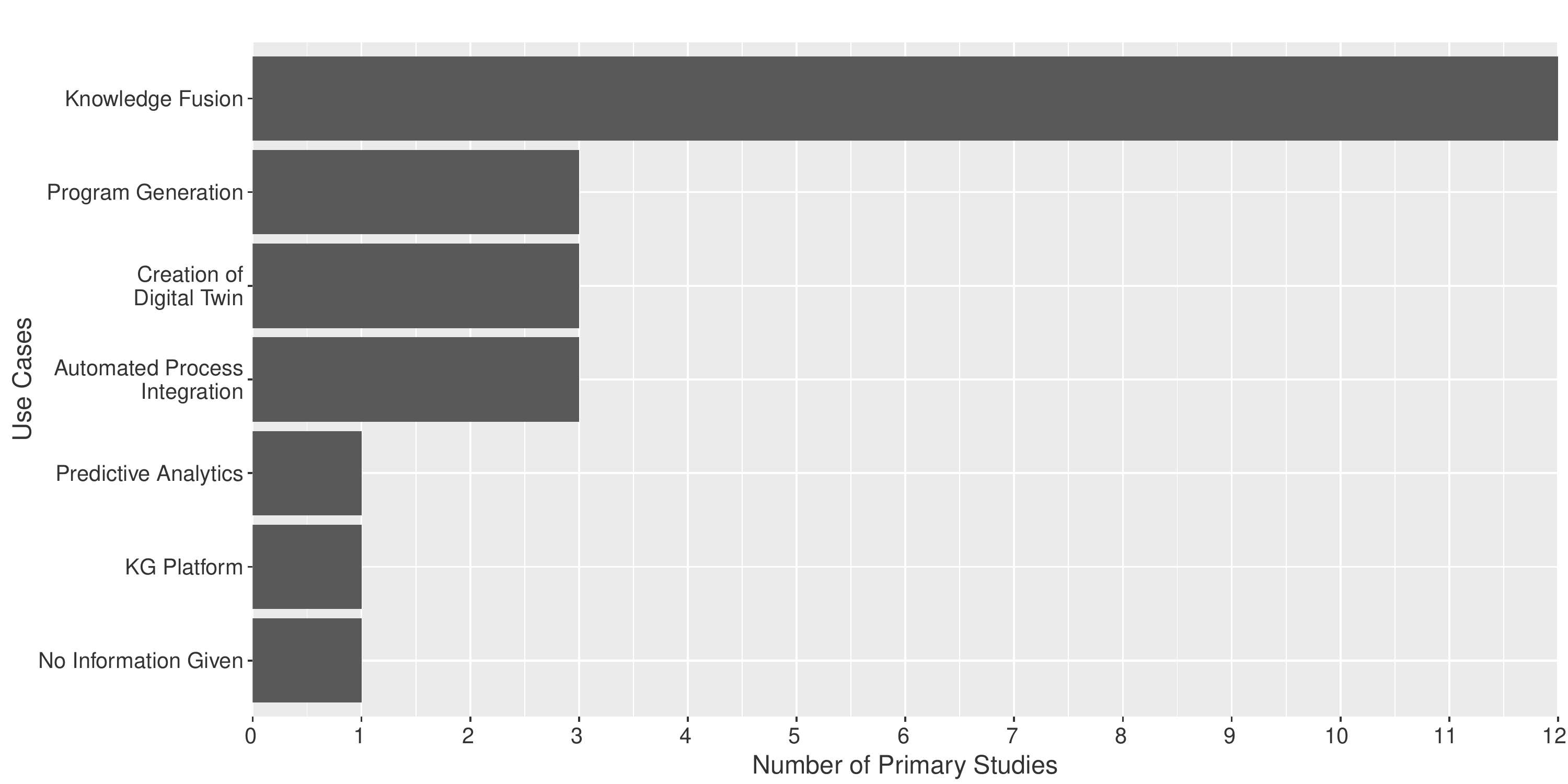}
\end{figure}

We have analyzed each primary study for the motivating use case, i.e., why a knowledge graph has been constructed. Figure \ref{FigUseCase} provides an overview which use cases have been described in the selected primary studies. 50~\% of all primary studies, i.e., 12 studies, describe a knowledge fusion use case where KGs are used for integrating data from different information sources with each other. 3 primary studies are using KGs for automatically integrating separated manufacturing processes with each other, 3 primary studies deal with the creation of digital twins based on data stored in a knowledge graph, 3 primary studies use KG data for the automated generation of source code. 1 primary study deals with the development of a general KG management platform, 1 primary study describes a predictive analytics use case. For one of the primary studies no concrete use case could be identified. Table \ref{UseCasesTable} links the assignment of each primary study to its motivating use case. 

\begin{table}[!htbp]
\centering
\renewcommand{\arraystretch}{1.3} \caption{Knowledge Graph Use Cases}
\label{UseCasesTable}
\centering
\begin{tabular}{@{\extracolsep{-5pt}} ll} 
\hline
\bfseries Use Case & \bfseries Studies \\\hline
Knowledge Fusion & \citeS{5} \citeS{22}  \citeS{11} \citeS{12}  \citeS{13} \citeS{2}  \citeS{9} \citeS{10}   \citeS{18}   \citeS{19} \citeS{21} \citeS{16}  \\
Creation of Digital Twin & \citeS{3} \citeS{4} \citeS{7}  \\
Automated Process Integration &  \citeS{23} \citeS{6} \citeS{15}   \\
Program Generation &  \citeS{14} \citeS{24} \citeS{20}  \\
Predictive Analytics & \citeS{1} \\
Knowledge Graph Platform & \citeS{17}  \\
No Information Provided & \citeS{8}  \\\hline
\end{tabular}
\end{table}

\begin{figure}[!htbp]
\centering
\caption{Knowledge Graph System Kinds}
\label{FigSystemKind}
\includegraphics[width=0.95\maxwidth]{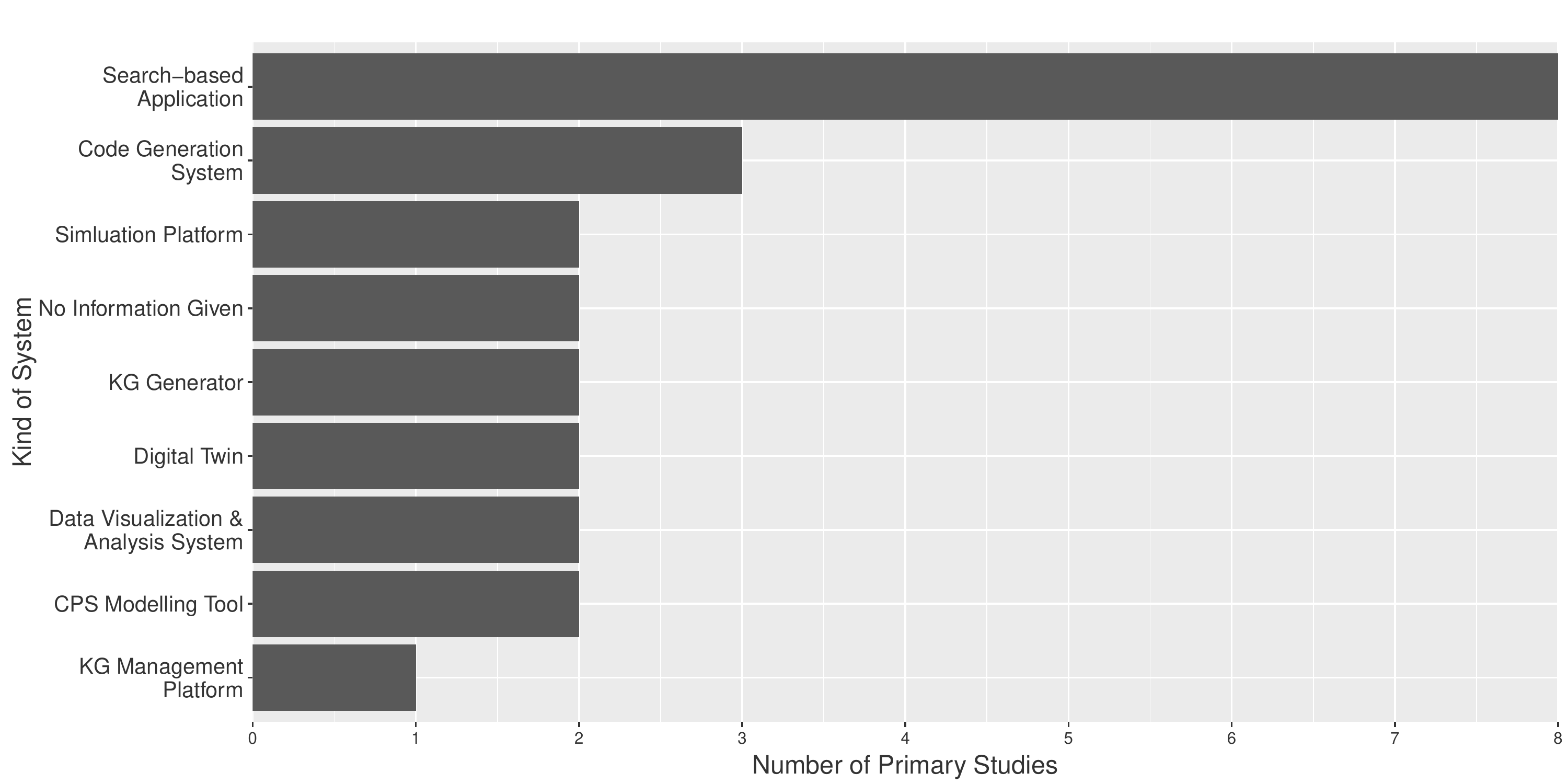}
\end{figure}

Finally, we have investigated which kinds of systems have been developed on top of a knowledge graph. Figure \ref{FigSystemKind} depicts the identified kinds of systems. In 8 cases a search-based application, i.e, an application providing a search engine based on semantic technologies, has been developed. In 3 cases, a code generation system has been proposed. 2 primary studies demonstrated the development of a data visualization and analysis system. In 2 cases a digital twin has been created, 2 studies supported the modelling/description of a cyber-physical system, and 2 studies provided a simulation platform. In 2 studies a system for the automated construction of a knowledge graph was proposed, 1 study proposed a KG management system. In 2 primary studies, no concrete system based on a knowledge graph was developed. Table \ref{SystemKindTable} lists which kinds of system have been described by each primary study. 

\begin{table}[!htbp]
\centering
\renewcommand{\arraystretch}{1.3} \caption{Knowledge Graph-based System Kinds}
\label{SystemKindTable}
\centering
\begin{tabular}{@{\extracolsep{-5pt}} ll} 
\hline
\bfseries Use Case & \bfseries Studies \\\hline
Search-based Application &  \citeS{5} \citeS{4} \citeS{13} \citeS{1} \citeS{2}     \citeS{6}   \citeS{18}  \citeS{21}    \\
Code Generation System  & \citeS{14} \citeS{24} \citeS{20}    \\
Digital Twin  & \citeS{3} \citeS{7}    \\
Data Visualization and Analysis System  & \citeS{22} \citeS{10}   \\
CPS Modelling Tool  & \citeS{11} \citeS{12}  \\
Simulation Platform  & \citeS{23} \citeS{15}    \\
Knowledge Graph Generator & \citeS{19}  \citeS{16}  \\
KG Management Platform & \citeS{17}  \\
No Information Provided &  \citeS{9}  \citeS{8} \\\hline
\end{tabular}
\end{table}

%Here we need to talk about which industry category each primary study belongs to as well as whether a use case has been provided and what topic each use case focuses on.

%Most works in this context include a case of use in the industrial domain. The great abundance of use cases in the literature, as opposed to more generalizable approaches, denotes that the research was initiated for fulfilling some kind of industrial needs. Most of the use cases belong to the category of Industry 4.0, although there are also several works that have focused on a specific company. The lack of open data in this context, only makes possible the validation of the proposals through proprietary data of the companies. This leads us to advise that if research in the field of manufacturing and production is to continue to be stimulated in the future, it will be necessary to have more and better open data sets of an industrial nature, so that researchers can validate their methods and tools. Otherwise, research will continue to take place in closed environments and be inspired by individual needs as has been the case up to now.
%\todo[inline]{The answers to the RQs partially already contain discussions. As a consequence some discussions are contained in Section 4, while some discussions are held in Section 5. }

%% file: 6_discussion.tex
\section{Discussion} \label{discussion}
In this section, we provide a discussion of the summarized results obtained from the classification and analysis of our study, along with an identification of gaps and opportunities for future research, and a discussion of the threats to validity of our study. %The section has been structured in the same way as the research questions, although it ends with general findings, open research challenges and the threats to validity. 

\subsection{About the bibliometric key facts of KG for publications (RQ1)}
The first point to highlight is that the application of knowledge graphs in manufacturing and production, although in an early stage, is attracting a lot of attention from both industry and academia. It should be noted that most primary studies that have been considered in this structured literature review come from the field of Computer Science. The Engineering and Business, Management, and Accounting sectors are far behind. This may be because the Computer Science community has led the way in knowledge representation models from the beginning. Today, knowledge graphs are understood as the natural evolution of such models to make them more adaptable to novel paradigms based on cloud and edge web platforms as well as new solutions for massive data and knowledge storage and maintenance. Furthermore, as can be observed in the results shown in the previous section, most of the studies selected have been published in conferences proceedings to date. This is often an indicator that the research topic is still emerging. In fact, in most fields of knowledge, as the state-of-the-art matures, more journal papers holding archival value are produced. Therefore, the case of knowledge graphs does seem to be no exception.

With regard to the countries leading research in knowledge graphs in the field of manufacturing and production, it could be observed that Germany and China are the two countries in which most primary studies originate from. Both countries also belong to the top manufacturing countries on a value-added basis, i.e., in 2018 China was the top manufacturing country and Germany held position 4\footnote{see https://www.statista.com/chart/20858/top-10-countries-by-share-of-global-manufacturing-output/}. Analyzing the relation between the origin countries of the primary studies and the leading manufacturing countries further reveals that the United States as second largest manufacturing nation only takes the fourth rank in the origin countries of the primary studies. For Japan as the third largest manufacturing country we could not find any primary study. Since the application of knowledge graphs in manufacturing and production is still an emerging topic (see above) and the number of primary studies is still low, the analysis of the origin countries should be seen as a current snapshot that can easily change in the future. 

\subsection{About the research  type  facets that the identified publications address (RQ2)}
\begin{figure}[!ht]
\centering
\caption{Summary of classification}
\label{FigOverview}
\includegraphics[width=1.0\maxwidth]{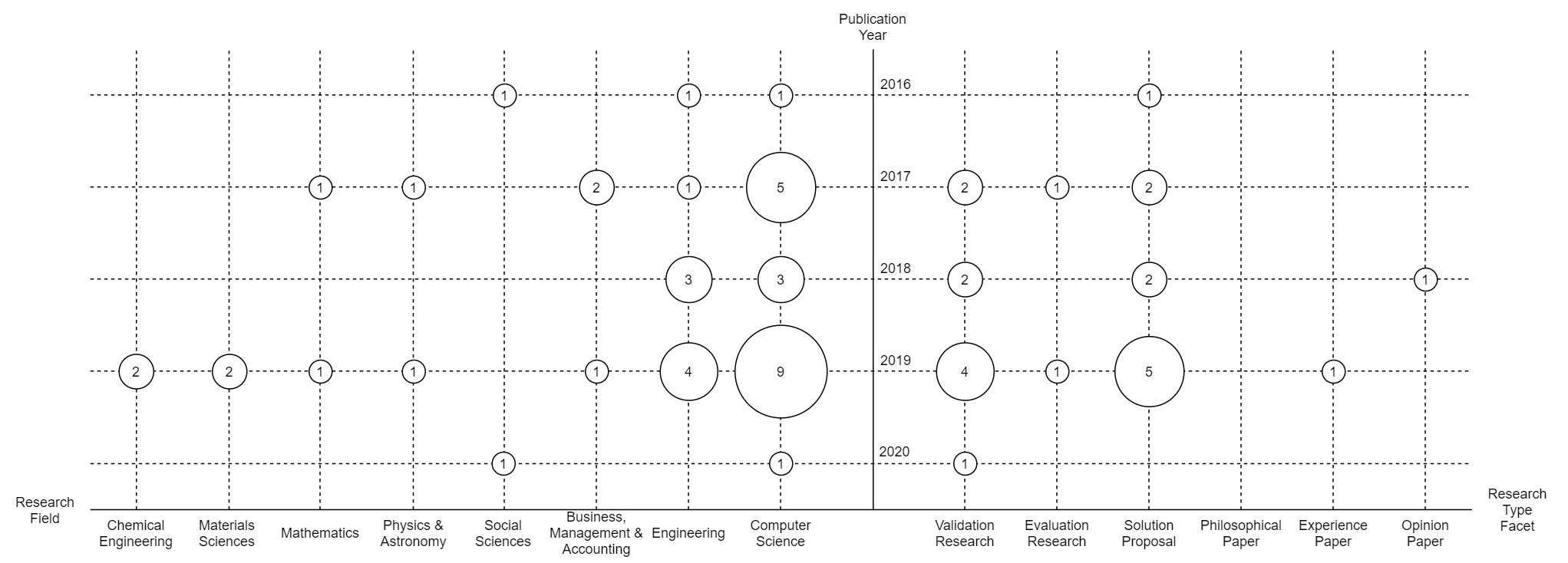}
\end{figure}

It is also noteworthy that the number of validation research papers which provide sound evidence of the usefulness of knowledge graphs in the production and manufacturing sector is still rather limited. So far most articles are proposing a solution for potential issues that can occur in production environments. However, these works have not been implemented in a real-world setting. Despite results obtained from validation research papers provide solid foundations for the employment of knowledge graphs in production and manufacturing companies and show the usefulness of such approaches in this field. Knowledge graphs in the field of manufacturing and production is a very important path that cannot be ignored if we take a look at the evolution of topics, the number of publications and research type forum over time as shown in Figure \ref{FigOverview}. At the same time, experience papers and opinion papers are in the minority at the moment. The fact that the new approaches based on deep learning have facilitated and accelerated the proposal of fully automated solutions does not seem to provide new views on their potential impact for production and manufacturing companies. This can probably change when advances are made to process and reason about numerical and tabular data, as these are the most frequent types of information in this domain.

\subsection{About the specific knowledge graph characteristics (RQ3)}
Another interesting takeaway is that against the vast majority of KG literature, knowledge graphs in an industry setting are usually constructed as a top-down approach and more property graphs are employed as compared to the general literature on knowledge graphs. Thus, it seems that in an industry setting top-down knowledge graphs enjoy higher popularity. This could be caused by the fact that a more closed-world approach fits better in a production setting as restrictions are set conditionally on the conducted task. This means that the manufacturing and production sector is still reluctant to incorporate the most recent advances in knowledge graph embeddings\footnote{Note: as mentioned in the Introduction, a KG embedding represents the data in the graph as numeric vectors.} to facilitate their analysis via machine learning and deep neural networks. 
%, i.e., the project of knowledge graphs into continuous vector spaces 
In this way, it seems that this sector is not largely profiting from the advanced techniques that allow working effectively and efficiently with embeddings to complete a quite extensive set of cognitive tasks. These tasks include the automatic prediction of new nodes and edges, the automatic link to other knowledge graphs of an analog nature, etc. However, there are a few innovative works that do not presuppose any structure beforehand and allow the knowledge graph to evolve freely, to discover facts or relationships that might be interesting. 

Last but not least and in relation to the previous point, it is worth mentioning that barely any paper with respect to this area is using machine learning or deep learning methods to expand the knowledge graph. We guess that this observation could be caused by the fact that companies want to know why decisions have been made and the missing explainability of 'black box' models such as neural networks is an uncertainty factor companies do not want to deal with in their production process. Additionally, as knowledge graphs mainly deal with textual pieces of information the number of machine learning tools that can be used for the prediction of categories is rather limited or have unrealistic assumptions to provide adequate results, at least if compared to situations that require the processing of numerical information, as is usually the case in the industrial sector.

In short, although the options in terms of representation and evolution of the knowledge graphs are diverse, most of the solutions proposed to date are eminently conservative and do not yet make intensive use of the new developments with regards to machine learning. We envision that, however, these advances will be increasingly incorporated into the existing body of literature soon as new solutions based on bottom-up knowledge graphs will be proposed in combination with more mature deep learning techniques that allow to be used in still unexplored yet relevant domains.

\subsection{About the application scenarios (RQ4)}

With regard to the application domain analysis which answers where knowledge graphs have been employed in manufacturing and production, it could be observed that for 50\% of the primary studies no explicit application domain could be determined. We see two reasons for this: First, many primary studies are still at an early stage of research (see also RQ2) and have not yet been evaluated in an industrial setting. Therefore, no concrete field of manufacturing could have been determined. Second, some primary studies propose general solutions that are not only applicable to a single field of manufacturing. For instance, analyzing data from production lines in a knowledge graph is a relevant topic for many manufacturing domains. 

With respect to the use cases for which knowledge graphs have been applied, it could be observed that knowledge fusion is the main use case for knowledge graph applications, which has been exploited by half of the analyzed studies. This comes at no surprise, since the integration of different data sources can be seen as a strength of knowledge graphs. Further, knowledge fusion can also be seen as a foundation/prerequisite for other use cases, i.e., the creation of digital twins, automated process integration, program generation, and predictive analytics. All those use cases build upon the integration and analysis of data from different data sources. The spectrum of use cases also shows the potential of knowledge graphs as underlying technology for production and manufacturing companies.

Analyzing the kinds of systems that have been developed in the primary studies revealed that in most cases (8 studies) search-based applications have been built based on knowledge graphs. These systems provide means for searching and exploring the data stored in a knowledge graph and allow analyzing relationships between data that has initially been stored in different data sources. Closely related to search-based applications are data visualization and analysis systems, which have been subject of research in 2 studies. All these systems focus on supporting analytical tasks and knowledge discovery. Code generation systems have been developed as part of 3 studies. In 2 studies, source code for autonomous robots/vehicles was generated, in 1 case test cases were generated from specifications. This shows the possibility of using knowledge graphs for automating constructive activities, i.e., software development tasks, which represents a paradigm shift from requirements-driven to data-driven development\cite{bosch2018takes}. Multiple system kinds deal with digital representations manufacturing and production processes. This includes the creation of digital twins, the automated modelling (documentation) of cyber-physical systems, and the simulation of processes. Finally, 3 studies focus on the construction and management of knowledge graphs, rather than on use cases where knowledge graphs are further used. 2 of these approaches investigate the automated construction of knowledge graphs from existing data. 1 study has proposed a platform for the creation and management of knowledge graphs. These works can be seen as foundation for facilitating the wider application of knowledge graphs in the future.

In summary, the analysis of the current application scenarios of knowledge graphs in manufacturing and production shows the wide applicability and potential of knowledge graphs as underlying technology and foundation for a variety of different use cases and different system kinds. The different use cases (RQ4), the current number of primary studies (RQ1), and the current maturity of research (RQ3) show that knowledge graphs are still an emerging research topic with multiple open research challenges.  

%It should also be noted that of the industrial categories into which our primary studies can be placed, there is one clear category that stands out from the others: General. This is because most of the work falls within areas such as the Internet of Things (IoT) or unspecified industry 4.0. Some fields that also stand out are Automotive, Metallurgy, Robotics, or Chemistry. However, most of the categories proposed in the NAICS taxonomy remain largely unexplored. We envision, and in fact it is one of the objectives of this study, that as the field of knowledge matures new proposals will appear that cover almost all categories of that taxonomy.\todo[inline]{This aspect is not discussed in the answers to the research questions in the previous section}

\subsection{Open Research Challenges} \label{challenges}
After examining the existing literature on knowledge graphs in the field of manufacturing and production, we have identified up to five open research challenges that have not been adequately addressed to date. The following is a description of each of these open research challenges (ORCs).

\subsubsection{ORC1: Proper handling of numeric and tabular data} \label{challenge1}
Today, almost all of the solutions presented in the literature are specifically designed to work with the information of an eminently textual nature. While it is true that this is an important type of information in  industry, it is not precisely the predominant kind in manufacturing and production environments that usually work with machines and equipment that produce data of a numerical and even tabular nature. Tabular data is the type of data that is frequently represented in comma separated values and is usually one of the most common input methods in industrial environments, since it allows to model a wide variety of data associated with temporal aspects (timestamps), spatial aspects (coordinates), etc. However, this type of file does not usually have information that allows to give them a meaning by themselves. For that reason, being able to integrate them into Knowledge Graphs can open a wide range of possibilities to help their processing in tasks such as data analysis, data integration and even knowledge discovery. In recent times, intensive research work has been carried out on the problem\footnote{https://www.cs.ox.ac.uk/isg/challenges/sem-tab/} and it is likely that these advances, when they become available, may represent a qualitative leap forward in this sector.

\subsubsection{ORC2: Further research in real-time knowledge graphs} \label{challenge2} A fact that seems widely assumed among researchers and practitioners is the suitability of knowledge graphs to properly deal with data, information, and knowledge of different nature that might arrive through different channels and sources. However, in the course of this study, we have not observed a great number of works related to the temporal aspect when processing knowledge graphs. The manufacturing industry, however, requires solutions that can operate satisfactorily in environments with significant time constraints as many of the processes are automated and require a high degree of synchronization between them. In fact, we are convinced that further research on issues related to the fast updating knowledge graphs in real-time as information is received through different channels and sources will be a great advance in this sector. We look forward to new proposals in this regard when basic research would facilitate it.

\subsubsection{ORC3: Automatic linkage with other publicly available knowledge graphs} \label{challenge3} One of the main characteristics that have made knowledge graphs popular is the possibility to grow based on identifying similar pivot points in other knowledge graphs that have been designed independently but have ended up being offered to the public, for example, on the web. However, it seems that all the research that has been published in relation to the manufacturing and production sector do not take into account this characteristic when designing and implementing solutions. On the contrary, it seems that the community is working with knowledge graphs contained in silos that solve a specific task without requiring connection to other publicly available graphs. This means that not all the capabilities that knowledge graphs can offer are being used to the fullest. Obviously, to stimulate more intensive research in this direction, it is necessary to have better solutions for the automatic alignment of entities between different knowledge graphs. In this sense, it should be noted that very good methods have recently appeared\footnote{https://paperswithcode.com/task/entity-alignment} and will soon be used with certain guarantees of success.

\subsubsection{ORC4: A higher degree of coverage of all manufacturing and production domains} \label{challenge4}
Based on the current state of evaluation (RQ2), although we have seen a great deal of research work in a number of categories in the manufacturing and production sector, the truth is that there is still a lack of validation of existing approaches in this context as well as many industrial categories that remain largely unexplored. For example, there is an abundance of generic solutions for Industry 4.0 or the Internet of Things, and even there are already some innovative solutions for categories that are important from the economic point of view such as the chemical and automotive sectors. However, we can see that most categories of the NAICS classification have not yet been explored from the point of view of basic and applied research on knowledge graphs to automate and/or improve the industrial processes that they currently implement (evaluation research). We think that this question is related to the incipient emergence of knowledge graphs in this sector. Therefore, it is expected that in the coming years more evaluation research will appear in each categories that make up the NAICS taxonomy.

\subsubsection{ORC5: Collection of best practices concerning the development of reference architectures for KG development} \label{challenge5} 
It also appears from our study that no work has been done in the direction of compiling and understanding what the best practices are in this sector. It remains a great future challenge to be able to identify which are the reference architectures regarding the design, implementation and exploitation of knowledge graphs in industrial and production environments. We believe that a compilation of best practices in this area can be truly beneficial in maintaining high standards of quality results, as well as, saving resources in the form of money and time when developing new systems or making changes to existing ones. The truth is that until now all the development have been carried out without clear guidelines that might promote the best outcomes. On the contrary, each team has developed its own solutions to the best of its knowledge. We believe that it could be highly beneficial to have a set of best practices that can standardize the development of knowledge graphs in the field of manufacturing and production. In this way, the domain can grow sustainably and reach a high degree of development that is beneficial to all stakeholders involved.

\subsection{Threats to Validity}

To mitigate threats to validity \cite{wohlin2012experimentation} regarding subjective measures, all selected primary studies were analyzed according to defined criteria by all three authors. Studies with conflicting votes were analyzed in detail and discussed in the group before they were re-evaluated again by all researchers. 

Exclusion of relevant studies during screening is another threat to validity. We tried to mitigate this by clearly defined criteria and a multi-stage screening process that facilitated in-depth analysis of studies with ambiguous evaluations. 

Our mitigation strategy against low statistical power is to use five academic libraries and Google Scholar as additional data source to obtain the most complete possible result set of publications. We also considered alternative terms for describing sub-fields in the field of manufacturing and production as keywords for the search process to mitigate risks of publication selection and instrumentation.

%% file: 7_conclusion.tex
\section{Concluding Remarks} \label{conclusion}
In this paper, we have presented a systematic literature review on knowledge graphs in manufacturing and production environments. 
With the growing amount of scientific literature on KGs, an overview on the current development is needed to assess the applicability for this specific domain. 
% As the research field of KGs and consequently, its corpus of associated literature grows, solutions are needed that can provide a complete vision of how such an area is developing. 
Thus, we have come to weave a systematic literature review of one of the most prominent fields of application for knowledge graphs. 
To date, a lot of attention has been paid to fundamental research in KGs as attested by the growing literature on knowledge graph surveys. %However, its application in the field of manufacturing lacks a systematic study.

After carrying out a systematic search for primary studies, we have classified the identified studies according to four facets: (1) bibliometric key facts, (2) research type, (3) knowledge graph characteristics, and (4) application scenarios. Based on this classification, we have analyzed the current state of research and identified open research challenges.

We identified that almost 90~\% of the research has been conducting in one of the top 9 manufacturing countries worldwide\footnote{See, \href{https://www.statista.com/statistics/456342/realtive-comparison-of-value-added-in-manufacturing-of-leading-countries/}{Statista}.}. This highlights the fact that KGs are of particular importance for the manufacturing and production domain. Furthermore, the majority of research papers have been published in conferences. This fact illustrates that this research area is still in an early stage. It should also be noted that more than 95~\% of all published studies underwent a review process and are thus academically sound. Another interesting fact is that almost all papers belong to the category of computer science or engineering. Barely any evaluation papers was identified and more than 80~\% of the selected studies are either solution proposals or validation research papers. Notably, almost 70~\% of all KGs in manufacturing and production follow a top-down approach whereas slightly more than 30~\% are constructed employing a bottom-up approach. Finally, 87.5~\% of the relevant KGs have been identified as RDF graphs whereas 12.5~\% belong to the category of property graphs.

The general trend that can be seen is that this area of knowledge, albeit upward, is still very fragmented. This means that several research groups, as well as public and private organizations, focus their research on small sections and manufacturing sectors that are of interest to them, but they do not usually look further beyond. In addition, most solutions are focused on the classical use of graph-oriented computing, so it does not take full advantage of new methods based on deep learning that allows the processing of graphs. Nevertheless, although timidly, some works are emerging and our forecast is that the number of works in this sense will shoot up in the near future. Furthermore, the research activity has not yet covered a good part of the categories included in the NAICS classification, but we believe that new works will appear that will cover practically all the categories. 

Although an upward trend can be discerned, we think that the use of knowledge graphs in manufacturing industries has not yet fully reached its peak. But the figures evidence that is currently and most likely will remain a very active field of research in the near future. After all, knowledge graphs offer an effective and efficient way to address some of the problems that have traditionally plagued the manufacturing and production environments. So as future work, we will continue to monitor the literature in this sector to gain a better understanding on how knowledge graphs can boost innovation and ensure that there are no remaining gaps to fill.

%% file: 8_acknowledgment.tex
\section*{Acknowledgment}

Dr. Georg Buchgeher, Dr. David Gabauer, and Dr. Jorge Martinez-Gil acknowledge the support from the Interreg Österreich-Bayern 2014-2020 Programme funded under grant agreement number \textit{(AB292)}. Furthermore, the research reported in this article has been partly funded by BMK, BMDW, and the Province of Upper Austria in the frame of the COMET Programme managed by FFG. Finally, the authors would like to thank \textit{(editor name)}, the editor in chief of the \textit{(journal)}, and the helpful comments of the anonymous referees which significant improved our manuscript. Any remaining errors are solely ours.

% Finally, the authors would like to thank \textit{(editor name)}, the editor in chief of the \textit{(journal)}, and the helpful comments of the anonymous referees, which significant improved our manuscript. Any remaining errors are solely ours.

%% file: 9_bibliography.tex
\begin{singlespace}
\bibliographystyle{apalike}
\bibliography{literature}

\bibliographystyleS{elsarticle-num}
\bibliographyS{literature}

\end{singlespace}